%% file: main.tex
\pdfoutput=1
\documentclass[11pt]{article}

\usepackage[preprint]{acl}

\usepackage{times}
\usepackage{latexsym}

\usepackage[T1]{fontenc}
\usepackage[utf8]{inputenc}

\usepackage{microtype}

\usepackage{inconsolata}

\usepackage{graphicx}

\usepackage{fancyhdr}
\usepackage{lastpage}
\fancypagestyle{firstpage}{
  \fancyhf{}
  
  \fancyfoot[C]{\emph{Findings of the Association for Computational Linguistics: EMNLP 2024}, pages 14642--14655\\
  November 12--16, 2024 \\ \textcopyright~2024 Association for Computational Linguistics}
}

\usepackage{mathtools,amssymb}
\usepackage{dsfont}
\usepackage{subcaption}

\usepackage[utf8]{inputenc} % allow utf-8 input
\usepackage[T1]{fontenc}    % use 8-bit T1 fonts
\usepackage{hyperref}       % hyperlinks
\usepackage{url}            % simple URL typesetting
\usepackage{booktabs}       % professional-quality tables
\usepackage{amsfonts}       % blackboard math symbols
\usepackage{nicefrac}       % compact symbols for 1/2, etc.
\usepackage{microtype}      % microtypography
\usepackage{xcolor}         % colors
\definecolor{bigbird_color}{RGB}{46,139,87}
\definecolor{longformer_color}{RGB}{107,143,179}
\definecolor{hipool_color}{RGB}{161,116,161}
\definecolor{hipool_bar}{RGB}{147,112,219}
\definecolor{longformer_bar}{RGB}{143,188,143}
\definecolor{bar_color}{RGB}{70,130,180}

\usepackage{tikz}
\usepackage{pgfplots}

\usepackage[normalem]{ulem}
\usepackage{multirow}
\usepackage{wrapfig}
\usepackage{algorithmic}
\usepackage{graphicx}
\usepackage{bm}
\usepackage{amsmath}
\usepackage{amssymb}
\usepackage{mathtools}
\usepackage{amsthm}
\newcommand{\xhdr}[1]{{\noindent\bfseries #1}.}
\newcommand{\mat}[1]{\mathbf{#1}}
\newcommand{\ve}[1]{\mathbf{#1}}
\newcommand{\ten}[1]{\bm{\mathcal{#1}}}

\theoremstyle{plain}
\newtheorem{theorem}{Theorem}[section]

\newtheorem{lemma}[theorem]{Lemma}

\theoremstyle{definition}

\theoremstyle{remark}

\usepackage{algorithm}
\usepackage{algorithmic}
\usepackage{colortbl}

\title{Long Sequence Modeling with Attention Tensorization: \\ From Sequence to Tensor Learning}

\author{
    Aosong Feng, Rex Ying \and Leandros Tassiulas \\
    Yale University\\
    \texttt{\{aosong.feng, rex.ying, leandros.tassiulas\}@yale.edu}
}

\begin{document}
\maketitle
\thispagestyle{firstpage}
\begin{abstract}
As the demand for processing extended textual data grows, the ability to handle long-range dependencies and maintain computational efficiency is more critical than ever.
One of the key issues for long-sequence modeling using attention-based model is the mismatch between the limited-range modeling power of full attention and the long-range token dependency in the input sequence.
In this work, we propose to scale up the attention receptive field by tensorizing long input sequences into compact tensor representations followed by attention on each transformed dimension.
The resulting Tensorized Attention can be adopted as efficient transformer backbones to extend input context length with improved memory and time efficiency.
We show that the proposed attention tensorization encodes token dependencies as a multi-hop attention process, and is equivalent to Kronecker decomposition of full attention.
Extensive experiments show that tensorized attention can be used to adapt pretrained LLMs with improved efficiency.
Notably, Llama-8B with tensorization is trained under 32,768 context length and can steadily extrapolate to 128k length during inference with $11\times$ speedup, compared to full attention with FlashAttention-2. Code is avilable in \url{https://github.com/asFeng/tensorized_attention.git}.
\end{abstract}

\section{Introduction}
In the swiftly advancing field of natural language processing (NLP), large language models (LLMs) like GPT-4 and Llama have emerged as vital tools,  demonstrating expertise in comprehending and producing human language.
These complex scenarios often involve context lengths longer than those LLMs were trained on, posing significant challenges to Transformer-based architectures in handling long sequences.
For example, LLMs are expected to read long paragraphs or books to answer questions while they are usually trained with a much smaller context length (like 8k for Llama-3 \cite{touvron2023llama}).

Scaling up LLMs for long sequences presents significant challenges related to limited attention windows during training and length extrapolation during inference.
First, transformers usually adopt a bounded context window in the training phase due to quadratic computation costs \cite{vaswani2017attention, tay2022efficient}.
To mitigate these costs and expand the training context window, existing works usually adopt attention approximation methods like sparse, low-rank, and softmax-free attention.
Despite their efficiency and scalability with relatively long sequences, these methods gain less popularity due to inefficient implementations and incompatibility with existing pre-trained LLMs.

Given such pretrained LLMs, researchers start to focus more on the length extrapolation challenge: leveraging the short-range modeling power acquired during pre-training to handle unseen long-range dependencies.
Addressing this challenge requires methods that take into account both efficiency and effectiveness perspectives.
One line of thought is to employ full attention with hardware-efficient implementations such as FlashAttention and quantization for efficiency, paired with positional extrapolation \cite{su2024roformer} or interpolation \cite{chen2023extending} for enhanced performance.
While full attention captures all potential correlations, it significantly increases running time, and many correlations in long sequences are in fact unnecessary and distracting.
The second line of approaches involves using segmented or sliding windows for efficiency, supplemented by additional recurrent memory modules \cite{bulatov2023scaling, wang2023augmenting} or global sinks \cite{xiao2023efficient, han2023lm} to integrate segments.
However, it remains difficult for these additional modules to retain past or global information as windows move and update, posing a persistent challenge in processing long sequences effectively.

\begin{figure}[hb]
  \vspace{-10pt}
  \centering \includegraphics[width=0.99\linewidth]{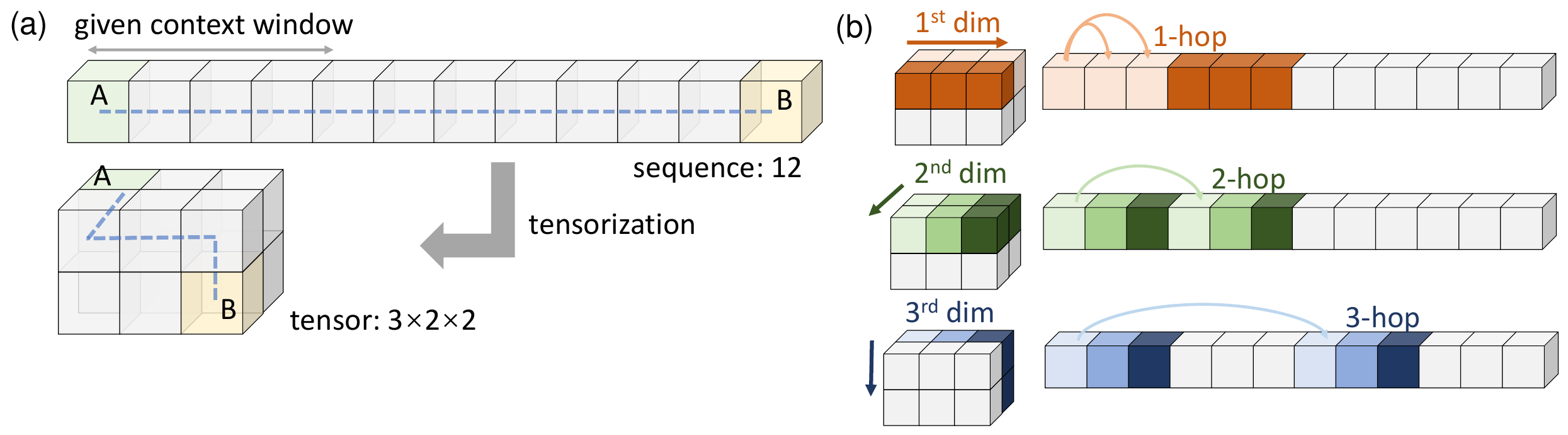}
  \vspace{-8pt}
  \caption{(a) The interaction distance from A to B is decreased from 12 in the sequence format to 5 in the tensor format and fits into context length. (b) Token interactions along each dimension are equivalent to multi-scale interaction in the original sequence.
  }
  
  \label{fig::tensor}
\end{figure}

\xhdr{Proposed Work}
To address these challenges in modeling long-range dependencies, we propose folding the one-dimensional long-range token interactions into a higher-order tensor, where each dimension can be effectively modeled with short-range interactions.
As shown in Figure \ref{fig::tensor}(a), given a limited context window, tensorization reduces the interaction distance between two tokens, converting previously out-of-window interactions into within-window interactions.
Inspired by the compact representation of the input tensor, we generalize input at the attention layer from the conventional sequence to higher-order tensor format and propose a supporting tensorized attention mechanism that replaces traditional vector operations with corresponding tensor operations.

Unlike traditional attention approximations, tensorized attention is implemented with an efficient custom Triton kernel and can be adapted from pre-trained LLMs through continued pretraining.
Besides, the multi-dimensional encoding of token interactions provides tensorized attention with natural advantages for length extrapolation.
Given finite attention window size or memory budget, tensorized attention captures correlations along different dimensions and incorporates hierarchical multi-hop interactions in the original sequence at multiple scales as shown in Figure \ref{fig::tensor}(b), which exponentially extends the effective context length and avoids the time consumption of using full attention.
Compared to segmented and sliding windows, tensorized attention hierarchically merges interactions from low to high levels and does not need additional global or recurrent modules to retrieve past information under exponential extrapolation speed.

We then integrates Tensorized Attentions into LLMs to improve the model scalability to longer-sequence inputs.
Experiments with diverse NLP datasets show that tensorized attention improves the performance and efficiency of existing pretrained models on long-sequence tasks after continued pretraining.
Notably, tensorized attention enables Llmama-3-8B training under 32,768 context length and can steadily extrapolate to 128k length during inference with $11\times$ speedup (compared to full attention with FlashAttention-2).
The code is at \url{https://github.com/asFeng/tensor_att}

\section{Related Works}

\subsection{Context Window Extension}
Efficient transformers have adopted various attention approximation including sparse and low-rank methods to extend the context length that can be processed by one attention layer.
For example, sparse Transformer \cite{child2019generating}, Longformer \cite{beltagy2020longformer}, Bigbird \cite{zaheer2020bigbird} and Diffusers \cite{feng2022diffuser} adopt combinations of random, window, and global sparse attention targeting the long-sequence scenario. 
Data-adaptive sparsity has also been considered in recent works including
Reformer \cite{kitaev2020reformer} BiFormer \cite{zhu2023biformer}, DAT \cite{xia2022vision}, and SparseViT \cite{chen2023sparsevit}.
The proposed tensorization is compatible with such sparisty masks.
Low-rank attention is based on the assumption that the attention matrix has an intrinsic low-rank structure.
Linformer \cite{wang2020linformer}, Performer \cite{katharopoulos2020transformers}, Linear Transformer \cite{katharopoulos2020transformers}, Synthesizer \cite{tay2021synthesizer}, and LRT \cite{winata2020lightweight} achieve efficient attention with linear complexity by projecting attention matrix to a lower dimension in the column vector space along the query or key dimension.
However, these low-rank assumptions are mostly based on the vector space by directly reducing q,k,v sequence length, ignoring the hierarchical structure embedded in the attention matrix.
Given the observed structure of attention matrices, we show that attention representations in the tensor space which is constructed by a set of hierarchically decomposed attention blocks can be better approximated with low-rank matrices.

\subsection{Length Extrapolation}
Given limited context length during training, length extrapolation aims to extend the model's comprehension to sequences beyond training context length. These extrapolation methods can be mainly summarized into three categories including position encoding variation, window based approach, and memory/global token augmented approach.

Position encoding improves the model's understanding of the sequential structure of input sequences. 
RoPE \cite{su2024roformer} deploys distinct rotary matrices to calculate scores between keys and queries using relative position information. While widely adopted in Llama and PaLM, it has been shown limited in zero-shot extrapolation setting \cite{press2021train}.
% To tackle this, ALiBi \cite{press2021train} assigns a predefined penalty score to improve extrapolation performance. 
Position Interpolation is adopted by \cite{chen2023extending}, Yarn \cite{peng2023yarn}, and Pose \cite{zhu2023pose}, which interpolate the position indices within the pretrained context and improves RoPE extrapolation performance after finetuning.
Our method adopts position encoding along each tensor dimension and can be combined with existing methods to extrapolate along a specific dimension. 

Window-based attention adopts the limited context length to model the window attention and slide the window to extrapolate.
For example, LM-Infitnite \cite{han2023lm} adopt a $\Lambda$ shape attention to bound token interactions to nearby tokens within the window.
% GrowLength \cite{jin2023growlength} progressively grows the context length during the training phase to process more tokens under the same running time. 
Such window based approach are usually combined with recurrent memory like Transformer-XL \cite{dai2019transformer}, RMT \cite{bulatov2023scaling}, \cite{chevalier2023adapting} or global/external memory like \cite{izacard2020leveraging}, StreamingLLM \cite{xiao2023efficient}, \cite{wu2022memorizing}.
Our tensorization approach generalizes window attention to different tensor dimensions which can hierarchically capture multi-hop relationships. With tensorization, out-of-window token pairs along one dimension can be in-window along another dimension.

\subsection{Tensor learning}
Tensor or multidimensional array has been introduced into deep learning for various applications.
Tensor representation of weights is commonly used to compress neural networks
For example, \cite{novikov2015tensorizing} considers a decomposed tensor format (TT) to represent the weight matrix in MLP which can take tensorized inputs. 
They are also adopted to explore efficient structure in CNN \cite{lebedev2014speeding, denton2014exploiting, kossaifi2019t,phan2020stable} and RNN \cite{yang2017tensor,pan2019compressing,su2020convolutional}.
In transformers, \cite{ren2022exploring} and \cite{pan2023reusing} use tensor decomposition to compress the transformer weights.
\cite{alman2023capture} considers modeling high-order interactions by Kronecker decomposing key vector.
The work that is most similar to ours is \cite{ma2019tensorized} which adopts Block-Term Tensor Decomposition to model the 3-way interactions between query, key, and value.
Compared to these works about tensorizing the neural network weights for compression,
we adopt tensorization as a method to efficiently compress the token interactions, corresponding to the hierarchical structure embedded in the attention matrix.

\section{Attention Tensorization}
\begin{figure*}[htp]
  \vspace{-1pt}
  \centering \includegraphics[width=1.0\linewidth]{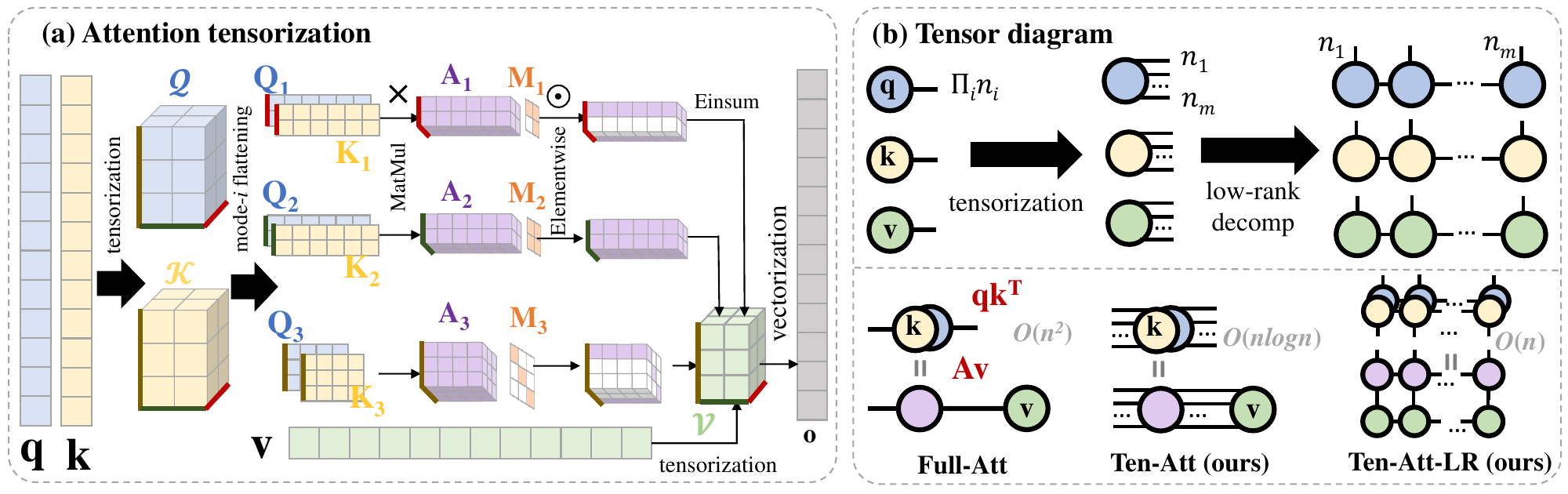}
  \caption{
  (a) Input sequences $\ve{q}, \ve{k}, \ve{v}$ are first tensorized into $\ten{Q}, \ten{K}, \ten{V}$.
  Each row in the middle represents the attention along one matching dimension of tensors, and all dimensions except the matching dimension of $\ten{Q}$ and $\ten{K}$ are flattened.
  The result from each row is used to sequentially update the value tensor $\ten{V}$.
  (b) Different types of attention processes can be visualized using a tensor diagram, where each circle represents data content and each edge represents a dimension.
  }
  \label{fig::model}
  \vspace{-5pt}
\end{figure*}

In this section, we first compare the representation of input sequences in both vector and tensor space.
To accommodate the tensorized input, we then introduce attention tensorization by changing conventional attention from the vector space to the corresponding tensor space.
Finally, we show that low-rank approximation of attention can be more efficient in tensor space compared to vector space, both empirically and theoretically.

\subsection{From Sequence to Tensor Learning}
In the following, we denote a scalar (order-0) with the lowercase letter $a$, a vector (order-1) with the lowercase bold letter $\ve{a}$, a matrix (order-2) with uppercase bold letter $\ve{A}$, and a tensor (order$>$2) with calligraphic bold letter $\ten{A}$. 
The order of input sequence is determined by the sequence length with feature dimensions omitted.
$\otimes$ denotes the outer product, Kronecker product, and tensor product (defined below) when operated on vectors, matrices, and tensors respectively. 
We use vector space to denote the space spanned by a set of base vectors and tensor space to denote the tensor direct product of several vector spaces.

\xhdr{Attention in Vector Space}
Given input sequence with $n$ tokens $\ve{x}\in\mathbb{R}^{n\times d}$, query $\ve{q}$, key $\ve{k}$ and value $\ve{v}$ are defined by linear projections as $\ve{q} = \mat{xW_q}, \ve{k} = \mat{xW_k}, \ve{v}=\mat{xW_v}$. The $n$-by-$n$ sparse attention matrix is then calculated as sampled scaled dot-product:
\begin{equation}
\label{eq::softmax}
\begin{split}
\mat{A}=\text{softmax}\left(\dfrac{\mat{qk}^\intercal}{\sqrt{d}} \circ \mat{M}\right), \ve{o} = \mat{A}\ve{v}
\end{split}
\end{equation}
where $\text{softmax}$ denotes the row-wise softmax normalization, and $\mat{M}$ is the attention mask. $\ve{o}$ is the output vector which is the updated value vector.
% In the multi-head setting, we allow different masks in different heads. %%*

Consider we have tensorized inputs $\ten{Q},\ten{K},\ten{V} \in \mathbb{R}^{n_1\times ...n_m}$ $\prod_{i=1}^{m}n_i=n$,  through order-$m$ tensorization which is achieved by reshaping $\ve{q}, \ve{k}, \ve{v}$ (discussed in Appendix \ref{asec::reshape}).
Our following tensor treatment is based on the principle that tensors interact with each other only through the same matched dimensions.
In other words, to update the $i$-th dimension of value tensor $\ten{V}$, we consider the interactions between the $i$-th dimension of $\ten{Q}$ and $\ten{K}$ while treating other dimensions as batch dimension.
The update of $\ten{V}$ can then be achieved by sequential updating from the first to the last dimension (irrelevant to the order of updating).

\subsection{Attention Tensorization}
To model the tensor interaction between $\ten{Q},\ten{K}, \ten{V}$, we propose attention tensorization which converts long-range interactions into short-range interactions along each dimension.
Since the length along each dimension is much smaller than the entire sequence, such tensorized attention can work with much less context window budget.
Besides, the hierarchical nature of the method enables exponential extrapolation to unseen long-sequences.
Attention process in Equation \ref{eq::softmax} can be tensorized as
\begin{equation}
\label{eq::tens_softmax}
\begin{split}
\ten{A}=\text{softmax}\left(\dfrac{\ten{Q}\otimes\ten{K}}{ \sqrt{d}} \circ \ten{M}\right), \ten{O} = \ten{A}\times\ten{V},
\end{split}
\end{equation}
where $\ten{A}$,  $\ten{M}$ $\in \mathbb{R}^{n_1\times n_1 \times ... n_m \times n_m}$, softmax is performed along every even dimension,  $\circ$ is still elementwise production, and $\ten{O}$ is the output tensor.
Following the aforementioned tensor interaction principle, $\otimes$ is the tensor (outer) product on the corresponding dimension, and can be written in Einsum notation as ``$n_1..n_i..n_m, p_1..p_i..p_m \rightarrow n_1 p_1 ..n_i p_i .. n_m p_m$'', or formally as
$(\ten{Q} \otimes \ten{K}) [..,j_{2i}, k_{2i+1} ,..]=\ten{Q}[.., j_i ,..]\ten{K}[.., k_i ,..]$ where $j_i$ represents the $j$-th element along the $i$-th dimension.
$\times$ is the tensor product between order-$2m$ and order-$m$ tensors, written in Einsum as 
``$n_1p_1..n_i p_i..n_m p_m, p_1..p_i..p_m\rightarrow n_1..n_i..n_m $''
or formally as $(\ten{A}\times \ten{V})[j_i] = \sum_{k_i=1}^{n_i} \ten{A}[..,j_{2i},k_{2i+1},..] \ten{V}[..,k_i,..]$.
The above calculations can be easily visualized using a tensor diagram as shown in Figure \ref{fig::model} (b).

\xhdr{Sequential Updating $\ten{A}\times\ten{V}$} 
Because the value tensor updating can be sequential which is irrespective of the updating order,
for efficient implementation, we adopt the sequential calculation of Equation \ref{eq::tens_softmax} and model the $i$-th dimension of $\ten{Q}, \ten{K}, \ten{V}$ interaction as 
\begin{equation}
\label{eq::tens_softmax_ithdimension}
\begin{split}
\ten{A}_i=\text{softmax}\left(\dfrac{\ten{Q}\otimes_i\ten{K}}{ \sqrt{d}} \circ \ten{M}_i\right), \ten{O} = \ten{A}_i \times_i \ten{V},
\end{split}
\end{equation}
where $\ten{A}_i$,  $\ten{M}_i$ $\in \mathbb{R}^{n_1\times .. n_i \times n_i..\times n_m}$, $\otimes_i$ and $\times_i$ are the tensor products confined to the $i$-th dimension with Einsum annotation as ``$n_1..n_i..n_m, p_1..p_i..p_m \rightarrow n_1 ..n_i p_i .. n_m$'' and ``$n_1..n_i p_i..n_m, p_1..p_i..p_m\rightarrow n_1..n_i..n_m $''.

We then express the this sequential updating in the matrix multiplication format using mode-$i$ flattening.
Mode-$i$ flattening reshapes a tensor $\ten{T}\in\mathbb{R}^{n_1\times ..\times n_m}$ into matrix $\mat{T}_i\in\mathbb{R}^{(n_1..n_{i-1}n_{i+1}..n_m) \times n_i}$ which can be interpreted as batching $n_1..n_{i-1}n_{i+1}..n_m$ mode-$i$ vectors.
We call the corresponding reverse process mode-$i$ folding.
Equation \ref{eq::tens_softmax_ithdimension} can be equivalently rewritten as
\begin{equation}
\label{eq::tens_softmax_ithdimension_mat}
\begin{split}
\mat{A}_i=\text{softmax}\left(\dfrac{\mat{Q}_i\mat{K}_i^\intercal}{ \sqrt{d}} \circ \mat{M}_i\right), \\
\mat{O} = \mat{A}_i  \mat{V}_i, \ten{O} = \text{fold}_i (\mat{O}_i),
\end{split}
\end{equation}
where $\text{fold}_i$ represents mode-$i$ folding, and it can also be shown that $\ten{A}_i = \text{fold}_i (\mat{A}_i)$.

We show how to use Equation \ref{eq::tens_softmax_ithdimension_mat} for sequential updates in Figure \ref{fig::model} (a).
For the $i$-th dimension update,  $\ten{Q},\ten{K}$ are matricized by mode-$i$ flattening.
The resulting batched attention matrix is then used to update the $i$-th dimension of the value tensor.
The computational complexity is decreased from $\mathcal{O}(n^2)$ to $\mathcal{O}(nlogn)$, and detailed complexity comparisons of each step are shown in Appendix \ref{asec::time_complexity}.
We summarize the forward pass of tensorized attention in Algorithm \ref{alg::forward}. 
For efficient calculation, we adopt Triton \cite{tillet2019triton} to implement the forward and backward process with hardware awareness.

\begin{algorithm}[tb]
   \caption{Tensorized attention forward}
   \label{alg::forward}
\begin{algorithmic}
   \STATE {\bfseries Input:} $\ten{Q},\ten{K}, \ten{V} \in \mathbb{R}^{n_1\times .. \times n_m}$
 
   \STATE Initialize $\ten{O} = \ten{V}$.
   \FOR{$i=0$ {\bfseries to} $m-1$}
   \STATE Mode-i flattening $\ten{Q},\ten{K}$ into $\mat{Q}_i,\mat{K}_i$
   \STATE $\mat{A}_i=\text{softmax}(\mat{Q}_i\mat{K}_i^\intercal/{ \sqrt{d}} \circ \mat{M}_i)$
   \STATE Mode-i flattening $\ten{O}$ into $\mat{O}_i$
   \STATE Value updates: $\mat{O}_i = \mat{A}_i \mat{O}_i$
   \STATE Mode-i folding $ \mat{O}_i$ into $\ten{O}$
   \ENDFOR
   
   \STATE Vectorize $\ten{O}$ to $\ve{o}$
   \STATE Return $\ve{o}$

\end{algorithmic}
\end{algorithm}

\begin{figure}[hb]
  \vspace{-10pt}
  \centering \includegraphics[width=0.99\linewidth]{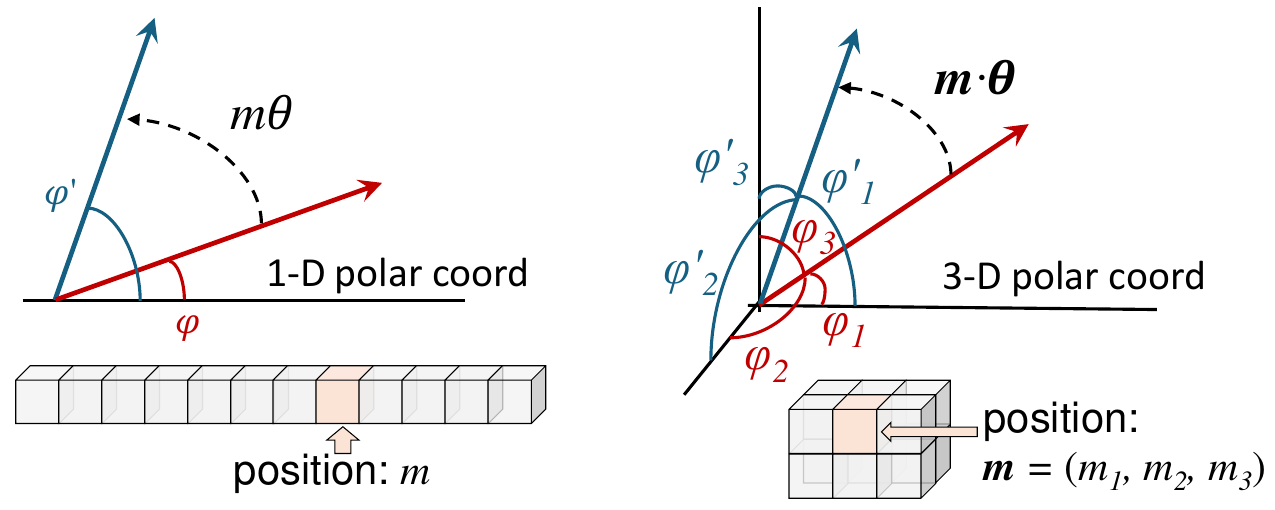}
  \vspace{-8pt}
  \caption{Comparison of token position in 1-D sequence and 3-D tensor under polar coordinates.
  }
  \label{fig::position}
\end{figure}

\xhdr{Tensorized Positional Encoding}
To indicate the position of a token in tensor, we convert the original sequential position into the corresponding hierarchical tensor position.
As shown in Figure \ref{fig::position}, the position of a token is defined as a vector that records the relative sequential position along each dimension.
Following pretrained Llama, we adopt ROPE to derive the corresponding positional encoding for sequential and tensorized position.
% (detailed in Appendix \ref{asec::rope}). 
Compared to sequential positioning, this hierarchical tensorized positioning enhances the extrapolation ability of LLMs.
Extrapolating the positional encoding along a specific dimension can exponentially extend the entire sequence, while keeping positional encodings on other dimensions within range.

\begin{figure*}[pbt]
  \vspace{-1pt}
  \centering \includegraphics[width=1.0\linewidth]{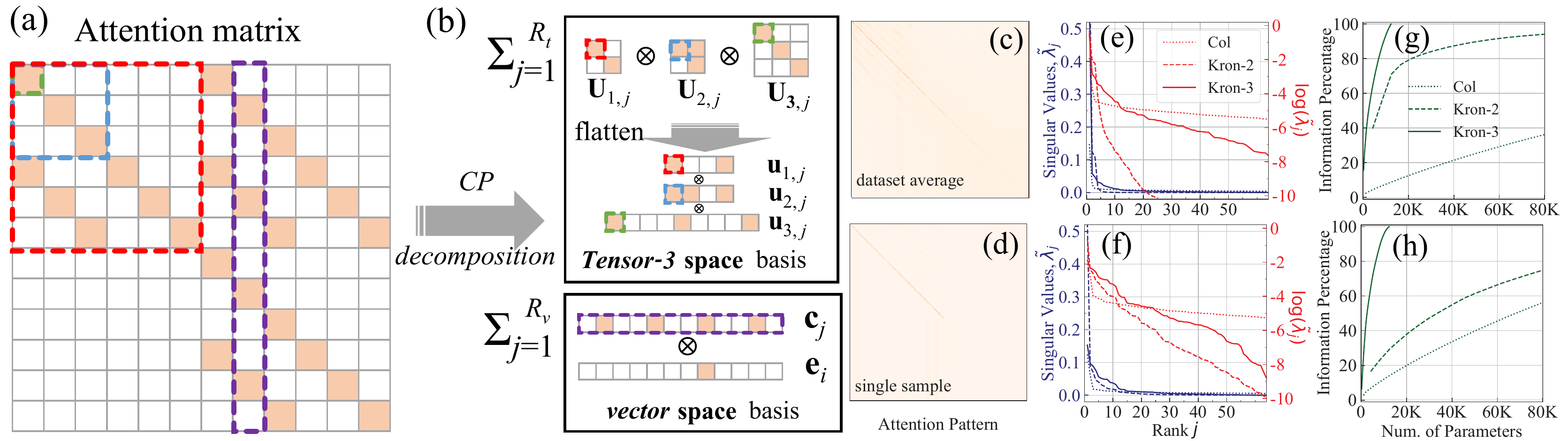}
  \caption{The attention or mask (a) can be decomposed to a set of columns or 2-D blocks which can be used to span vector or tensor space. (b) The total number of needed singular vectors to reconstruct the given example pattern $R_t=1<R_v=12$, shows the advantage of using tensor space for diagonal and structured patterns.
  (c,d) CP decomposition of real attention patterns (layer 3, head 2) from (c) dataset average or (d) single-sample pattern. 
  (e, f) Calculated spectrum as sorted normalized singular values $\Tilde{\lambda}_i$. 
  (g, h) Total percentage of information contained v.s. number of parameters needed for approximation.
  %%% change to (a)(b))(c)
  }
  \label{fig::tensor_space}
\end{figure*}

\subsection{Efficient Approximation in Tensor Space}
\label{sec::spectrum}
We focus on the properties of the attention matrix and show that the representation of attention in the tensor space can better capture the intrinsic hierarchical and low-rank structure embedded in data, compared to vector space.
By comparing the low-rank approximation performance, we show that 
to recover the same amount of information from the attention matrix, using representation in the tensor space requires much fewer parameters than vector space.

\xhdr{Attention Rank in Vector Space}
Given a 2-D matrix $\mat{A}\in\mathbb{R}^{n\times n}$ with columns $[\ve{c}_1, \ve{c}_2, ..., \ve{c}_n]$, the corresponding vector space is constructed as $ \ve{a}=\sum_{j=1}^{n}\lambda_j \ve{c}_j$ with $\lambda_j$ as a scalar. 
Alternatively, we take an equivalent definition of vector space as 
$\mat{A}=\sum_{j=1}^{n}\lambda_j\ve{c}_j\otimes\ve{e}_j$ where unit vector $\ve{e}_j$ has all $0$ except $1$ in $j$-th position.
The rank $r$ of the matrix $\mat{A}$ is then defined as the minimum $n$.
Since $\{e_j\}$ are linearly independent, the basis of vector space is $\{ \ve{c}_j \otimes \ve{e}_j \}_{j=1}^{r}$ or equivalently $\{ \ve{c}_j \}_{j=1}^{r}$.

\xhdr{Attention Rank in Tensor Space} We then generalize the above formulations to higher-order attention tensor $\ten{T} \in \mathbb{R}^{n_1\times...n_m}$ with order $m$. 
A rank-$r$ tensor $\ten{T}$ can be decomposed into the sum of $r$ rank-$1$ simple tensors using CP decomposition: $\ten{T}=\sum_{j=1}^{r}\lambda_j\ve{u}_{1,j}\otimes...\otimes\ve{u}_{m,j}=\sum_{j=1}^{r}\lambda_j\otimes_{i=1}^{m}\ve{u}_{i,j}$, with basis $\{ \otimes_{i=1}^{m}\ve{u}_{i,j} \}_{j=1}^r$.
For $n$-by-$n$ attention matrix $\mat{A}$ with corresponding tensor form $\ten{A}\in\mathbb{R}^{n1^2..\times n_m^2}$, we show that attention tensorization hierarchically encodes different levels of structural information in attention matrix into each dimension of $\ten{A}$.
As shown in Figure \ref{fig::tensor_space} (a,b), vectors in the tensor basis of $\ten{A}$ induced by attention tensorization are formed by hierarchical Kronecker decomposition of the original attention matrix.
More specifically, $\mat{A}$ is first decomposed to several block matrices through matrix Kronecker decomposition $\mat{A}=\sum_{j=1}^{r}\lambda_j\otimes_{i=1}^{m}\mat{U}_{i,j}$.
Then $\mat{U}_{i,j}\in\mathbb{R}^{n_i\times n_i}$ is flattened to $\ve{u}_{i,j}\in\mathbb{R}^{n_i^2}$.
The corresponding tensor space is then spanned by a set of ($r$ linearly independent) base tensors $\otimes_{i=1}^{m}\ve{u}_{i,j}$ with the representation of  $\ten{A}$ written as $\ten{A}=\sum_{j=1}^{r}\lambda_j\otimes_{i=1}^{m}\ve{u}_{i,j}$ with singular values $\lambda_{j}$.

\xhdr{Spectrum Analysis} We analyze and compare the spectrum of the attention matrix in both vector and tensor space.
Empirical attention statistics are obtained from pretrained RoBERTa with IMDB dataset.
We consider the tensorization from 512 seuqnece to $\{32, 16\}$ and $\{8, 8, 8\}$ tensor and denote the resulting space as tensor-2 and tensor-3 space, respectively.
To compare efficiency, we also calculate the number of parameters needed for approximation as the total number of singular vectors in use multiplied by the size of each singular vector $\otimes_{i=1}^{m}\ve{u}_{i,j}$. 
For example, approximating attention with a rank-3 vector in tensor-3 space needs $3\times (8^2+8^2+8^2)$ parameters.

As shown in Figure \ref{fig::tensor_space} (e-h), singular values in tensor space decay faster than those in vector space, and thus require fewer singular vectors and parameters to recover the same amount of information of $\mat{A}$.
For tensor space with different orders, $m=2$ needs fewer singular vectors than $m=3$ for approximation, but $m=3$ requires much fewer parameters than $m=2$ because the size of each singular vector is much smaller (e.g., $8^2+8^2+8^2 < 32^2+16^2$).
We defer similar discussions of the vision model to Appendix \ref{asec::nlp_spec}.
Intuitively, if the attention matrix is more hierarchically structured with respect to base 2-D blocks, it can be more easily (with a lower rank) decomposed into a low-order tensor space. Conversely, if the matrix exhibits a more irregular and complex pattern, it may be necessary to decompose it into a higher-order tensor space.
The following theorem demonstrates low-rank attention properties in tensor space, with proof in Appendix \ref{asec::theo}. 

\begin{theorem}
\label{thm::spectrum} For attention matrix $\mat{A}\in\mathbb{R}^{n\times n}$ and any column $\ve{y}\in\mathbb{R}^{n}$ of value vector along the feature dimension, there exists a low-rank matrix $\tilde{\mat{A}}\in\mathbb{R}^{n\times n}$ with rank 
$\mathcal{O}(3^{m}\text{log}_{2m}n)$
defined in tensor-$m$ space, such that $Pr\left(||\tilde{\mat{A}}\ve{y} -\mat{A}\ve{y} ||<\epsilon||\mat{A}\ve{y} ||\right)>1-\mathcal{O}(1)$, for any $\epsilon$.
\end{theorem}

\section{Experiments}
In this section, we conduct various experiments to demonstrate that tensorized attention can serve as an efficient backbone for LLMs in long-sequence applications. We first discuss the continued pretraining process, followed by an evaluation of its performance on downstream tasks and its extrapolation capabilities.
Additionally, we showcase the strong performance of tensorized attention in conventional NLP and time series models, as detailed in the Appendix \ref{asec::experiment}.

\subsection{Efficient Language Modeling Backbone}
We adopt three recent LLM backbones: OpenLlama-3B, Mistral-7B, and LLama-8B with context lengths 2,048, 8,192, and 8,192, respectively.
Based on their pretrained checkpoints, we then adopt continued pretraining using (1) tensorized attention with tensorized positional encoding (denoted as "-Tens") and (2) full attention with position interpolation (PI) (denoted as "-Full") as baseline.
We perform the pretraining on sampled RedPajama dataset \cite{together2023redpajama} which contains 20\% Arxiv 20\% Book and 10\% other data with data shorter than 2K words excluded.
For tensorized attention, we set the tensor dimensions to be $\{32,32,32\}$ corresponding to the training context length 32,768, while for full attention we adopt FlashAttention-2 for efficient implementation.
The continued pretraining is running for 20B tokens for both models.

\subsection{Pretraining Perplexity} 
\begin{figure}[ht]
  \vspace{-10pt}
  \centering \includegraphics[width=1\linewidth]{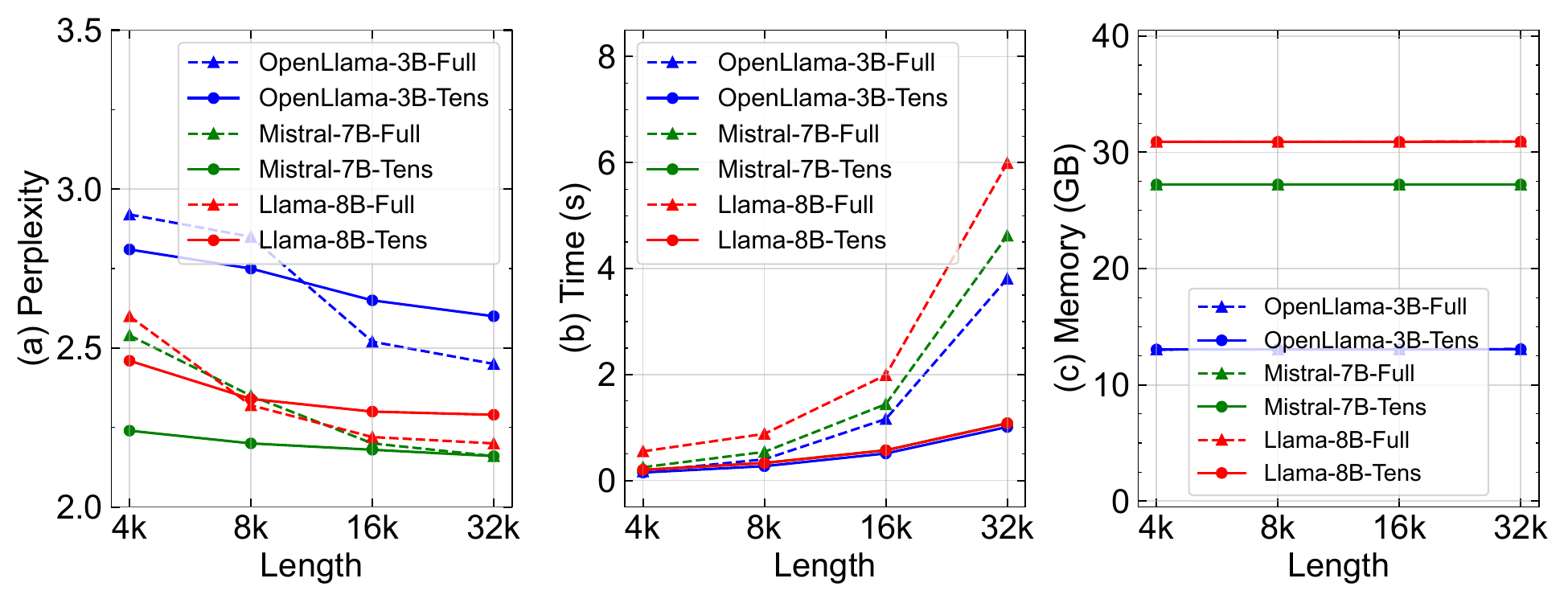}
  \caption{(a) Perplexity on Proof-pile test dataset after the continued pretraining. Comparison of GPU memory usage (b) and running time (c) efficiency for full and tensorized attention. The time and memory are calculated by averaging 50 forward passes and batch size 5.
  }
  \vspace{-10pt}
  \label{fig::ppl_flops}
\end{figure}

We evaluate the continued pretraining performance using next-token prediction perplexity on Proof-pile test datasets \cite{rae2019compressive}.
The perplexity and efficiency with different context lengths are shown in Figure \ref{fig::ppl_flops}.
As depicted in the figure, the model perplexity benefits from increased context length from 4k to 32k.
Compared to the standard full attention pretraining, tensorized attention can effectively extend the training context window to 32k while keeping low perplexity.
FlashAttention-2 can make memory usage steady by fixing block size after reaching the threshold, at the cost of increasing inference time.
Compared to FlashAttention, our kernel implementation for tensorized attention achieves consistently lower running times with similar memory usage, with the advantage becoming more pronounced for longer sequences.
It should be noted that the memory usage for the tensorization kernel and FlashAttention are nearly identical due to the blockwise attention calculation.

\begin{table*}[pt]
% \vspace{-5pt}
\centering
\resizebox{2.0\columnwidth}{!}{
\begin{tabular}{l|cccc|cc|cc|c} 
\toprule
\multicolumn{1}{c|}{\multirow{2}{*}{Model}}         & \multicolumn{4}{c|}{Common Sense}                             & \multicolumn{2}{c|}{World Knowledge} & \multicolumn{2}{c|}{Reading}  & \multirow{2}{*}{MMLU}  \\
\multicolumn{1}{c|}{}                               & HellaSwag     & SIQA          & WinoGrande    & PIQA          & NQ            & TQA                  & RACE-m        & RACE-h        &                        \\ 
\hline
Llama-7B                                            & 76.5          & 49.0          & 69.2          & 79.0          & 12.5          & 56.6                 & 41.2          & 39.5          & 46.1                   \\
Falcon-7B                                           & 76.3          & 49.1          & 67.1          & 80.5          & 14.6          & 50.9                 & 42.3          & 37.2          & 30.5                   \\
Gemma-7B                                            & 80.3          & 50.7          & 72.0          & 80.9          & 15.2          & 57.5                 & 43.6          & 38.3          & 63.8                   \\
StableLM-7B                                         & 71.7          & 44.1          & 69.1          & 79.8          & 9.1           & 46.4                 & 41.2          & 38.9          & 45.6                   \\ 
\hline
OpenLlama-3B-Full                                   & 65.2          & 44.8          & \textbf{63.5} & 76.2          & 6.3           & 31.5                 & 40.6          & 36.8          & 27.1                   \\
\rowcolor[RGB]{220,220,220} OpenLlama-3B-Tens & \textbf{69.4} & \textbf{45.3} & 63.2          & \textbf{78.0} & \textbf{7.2}  & \textbf{32.2}        & \textbf{42.3} & \textbf{37.7} & \textbf{27.4}          \\ 
\hline
Mistral-7B-Full                                     & 79.5          & 47.0          & \textbf{73.5} & 81.1          & 15.5          & 58.2                 & 45.2          & 39.8          & 62.5                   \\
\rowcolor[RGB]{220,220,220} Mistral-7B-Tens   & \textbf{80.5} & \textbf{49.1} & 73.2          & \textbf{81.3} & \textbf{15.8} & \textbf{58.8}        & \textbf{45.8} & \textbf{41.3} & \textbf{63.3}          \\ 
\hline
Llama-8B-Full                                       & \textbf{81.4} & 49.5          & 73.8          & \textbf{81.5} & 16.0          & 57.8                 & 44.8          & 40.3          & 65.4                   \\
\rowcolor[RGB]{220,220,220} Llama-8B-Tens     & 81.0          & \textbf{50.3} & \textbf{74.6} & 81.0          & \textbf{16.3} & \textbf{58.2}        & \textbf{45.2} & \textbf{42.6} & \textbf{66.2}          \\
\bottomrule
\end{tabular}
}
\caption{Downstream tasks from different domains including common sense, world knowledge, and reading comprehension. All benchmark models are running with FlashAttention-2.}
% \vspace{-5pt}
\label{tab::zero_shot}
\end{table*}

\begin{table*}[pt]
% \vspace{-5pt}
\centering
\resizebox{1.8\columnwidth}{!}{
\begin{tabular}{l|cc|cc|ccc|c} 
\toprule
\multirow{2}{*}{Model}                              & \multicolumn{2}{c|}{Multi-DocQA} & \multicolumn{2}{c|}{Summarization} & \multicolumn{3}{c|}{Few-shot}                                                                                                & \multirow{3}{*}{Time}  \\
                                                    & HotpotQA      & 2Wiki            & GovReport     & QMSum              & TREC          & TriviaQA                                                                                     & SAMSum        &                        \\ 
\cline{1-8}
AvgLength                                           & 9,151         & 4,887            & 8,734         & 10,614             & 5,177         & 8,209                                                                                        & 6,258         &                        \\ 
\hline
OpenLlama-3B-Full                                   & 18.4          & 20.1             & 12.4          & 15.6               & 52.3          & 72.7                                                                                         & 33.8          & 1$\times$              \\
\rowcolor[rgb]{0.863,0.863,0.863} OpenLlama-3B-Tens & \textbf{19.5} & \textbf{21.5}    & \textbf{13.5} & \textbf{16.2}      & \textbf{53.0} & \begin{tabular}[c]{@{}>{\cellcolor[rgb]{0.863,0.863,0.863}}c@{}}\textbf{74.8}\\\end{tabular} & \textbf{38.2} & 0.68$\times$           \\ 
\hline
Mistral-7B-Full                                     & 34.9          & 29.9             & 24.2          & 22.2               & \textbf{68.2} & \textbf{87.7}                                                                                & 41.3          & 1$\times$              \\
\rowcolor[rgb]{0.863,0.863,0.863} Mistral-7B-Tens   & \textbf{35.5} & 28.4             & \textbf{24.8} & \textbf{23.5}      & 68.0          & 87.5                                                                                         & \textbf{44.3} & 0.61$\times$           \\ 
\hline
Llama-8B-Full                                       & \textbf{45.3} & \textbf{34.5}    & 30.7          & \textbf{25.8}      & 71.1          & 86.1                                                                                         & 42.5          & 1$\times$              \\
\rowcolor[rgb]{0.863,0.863,0.863} Llama-8B-Tens     & 45.2          & 33.8             & \textbf{32.5} & 24.9               & \textbf{72.5} & \textbf{88.2}                                                                                & \textbf{44.0} & 0.73$\times$           \\
\bottomrule
\end{tabular}
}
\caption{Performance comparison of full and tensorized attention on LongBench.}

\label{tab::longbench}
\end{table*}

\subsection{Downstream Evaluation}
We follow Llama to report the zero- and few-shot results on various tasks after the continued pretraining stage.
Following the setting in \citet{eval-harness}, we adopt tasks include HellaSwag \cite{zellers2019hellaswag}, SIQA \cite{sap2019socialiqa}, PIQA \cite{bisk2020piqa}, and WinoGrande \cite{sakaguchi2021winogrande} for common sense reasoning ability with zero-shot evaluations;  NaturalQuestions (NQ) \cite{kwiatkowski2019natural} and TriviaQA (TQA) \cite{joshi2017triviaqa} for world knowledge understanding with zero-shot evaluations;
RACE-middle (RACE-m) and RACE-hard (RACE-h) \cite{lai2017race} for reading comprehension capabilities with zero-shot evaluations;
Massive multilingual language understanding (MMLU) \cite{hendrycks2020measuring} with 5-shot evaluations.
Table \ref{tab::zero_shot} indicates the pretrained tensorized attention outperforms both 3B, 7B and 8B benchmarks for world knowledge and reading tasks and achieves comparable results in common sense tasks.
To further evaluate the model performance on tasks with longer sequences, we follow the evaluation protocol in LongBench \cite{bai2023longbench} and show the task performance in Table \ref{tab::longbench}.
Tensorized attention consistently outperforms full attention in OpenLlama-3B with original context window 2,048 and achieves comparable results on Mistral-7B and Llama-8B with original context window 8,192, which shows the continued pertaining effectively extends the original context window using tensorized inputs.
Besides, tensorization enjoys less running time compared to full attention benchmarks (implemented with FlashAttention-2). For example, tensorized attention achieves $0.61 \times$ running time average across all tasks compared to full attention.

\subsection{Length Extrapolation}
LLM extrapolation techniques extend the model's comprehension to sequences beyond its initially observed length within the training context window.
Besides position- and window-based extrapolation, tensorization provides a new perspective to extend the context length by extrapolating the tensor on each dimension instead of the original sequence.
By increasing the context length by $p$ tokens on the $i$-th order, the total effective context length will increase by $p\cdot \Pi_{j\neq i} n_j$, which amplifies small increment on one dimension and effectively extends the total sequence length.
For example, for 32,768 tokens with $\{32,32,32\}$ tensorization, $1$ token increment on the first dimension is equivalent to 1,024 token length expansion in total. We extrapolate from the last (lower) to the first (higher) dimension for hierarchically encoding unseen positions.
% Compared to conventional vector extrapolation, tensor extrapolation avoids the explosive length growth in positional encoding and can handle exponential context expansion.

\begin{figure}[h]
  \vspace{-2pt}
  \centering \includegraphics[width=1\linewidth]{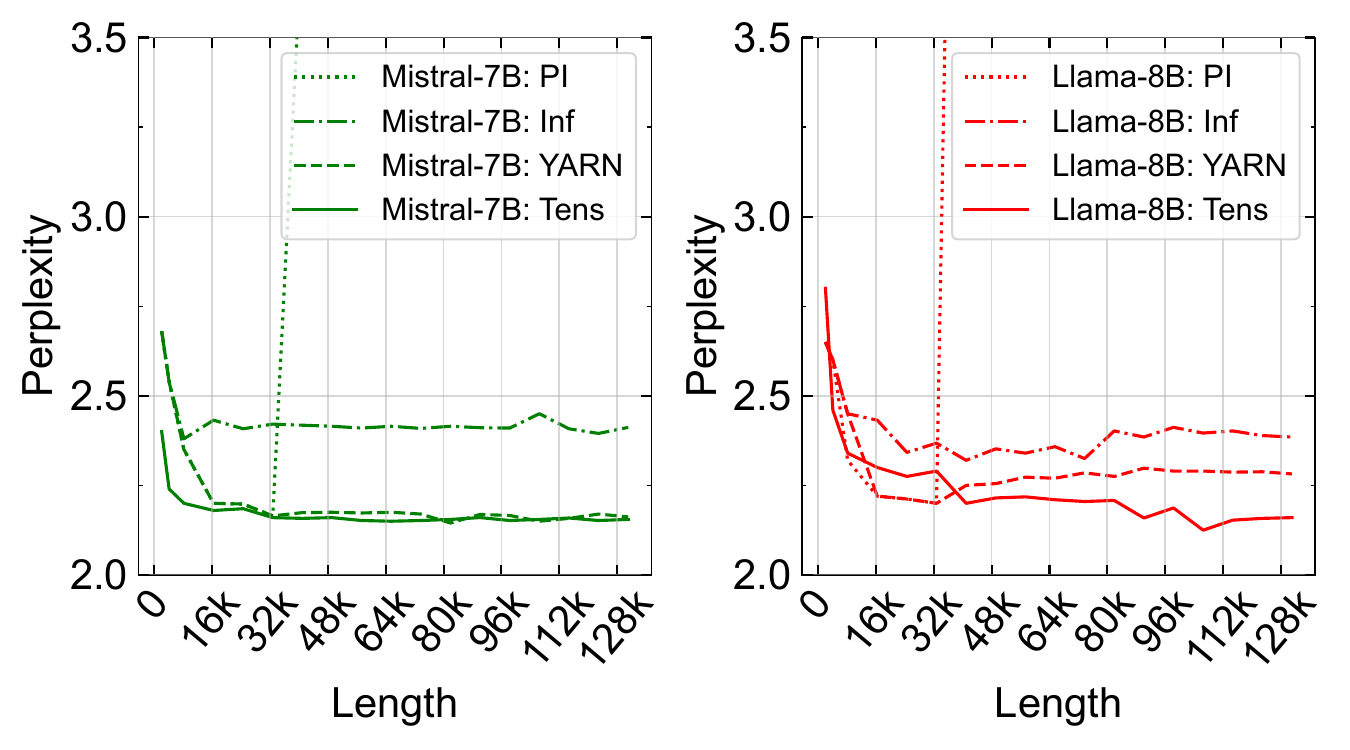}
  \caption{ The perplexity spectrum up to length 128k with different extrapolation methods.
  }
\vspace{-1pt}
  \label{fig::spectrum}
\end{figure}

We show the extrapolation performance and efficiency of tensorization using the perplexity of the Proof-pile test dataset in Table \ref{tab::extrapo} with spectrum in Figure \ref{fig::spectrum}. 
For comparison, we compare three widely used extrapolation methods: positional interpolation (PI) \cite{chen2023extending}, LM-Infinite (Inf) \cite{han2023lm} with attention distance 1,024, and YARN \cite{peng2023yarn} with scale factor 32.
While the original full attention model with positional encoding fails when surpassing the training context length, Inf, YARN, and Tens maintain performance from 32k to 128k, with Tens achieving consistently lower perplexity as sequence length increases, especially in the Llama-8B model. 
Additionally, the time improvement from attention tensorization becomes more pronounced with increasing context length. Notably, tensorized attention reduces the running time by $0.09 \times$ at 128k length compared to full attention for Llama-3 model.

\begin{table}[h]
\vspace{-5pt}
\centering
\resizebox{1\columnwidth}{!}{
\begin{tabular}{l|l|ccccc} 
\toprule
\multicolumn{2}{c|}{Length}                                              & 16k                                      & 32k                                      & 64k                                      & 96k                                      & 128k                                      \\ 
\midrule
\multirow{5}{*}{OpenLlama-3B} & PI                                       & 2.52                                     & 2.45                                     & -                                        & -                                        & -                                         \\
                              & InfLM                                    & 2.43                                     & 2.50                                     & 2.58                                     & 2.55                                     & 2.60                                      \\
                              & YARN                                     & 2.52                                     & 2.48                                     & 2.65                                     & 2.71                                     & 2.68                                      \\
                              & {\cellcolor[rgb]{0.863,0.863,0.863}}Tens & {\cellcolor[rgb]{0.863,0.863,0.863}}2.65 & {\cellcolor[rgb]{0.863,0.863,0.863}}2.60 & {\cellcolor[rgb]{0.863,0.863,0.863}}2.50 & {\cellcolor[rgb]{0.863,0.863,0.863}}2.53 & {\cellcolor[rgb]{0.863,0.863,0.863}}2.57  \\
                              & Time                                     & 0.44$\times$                             & 0.27$\times$                             & 0.14$\times$                             & 0.10$\times$                             & 0.08$\times$                              \\ 
\midrule
\multirow{5}{*}{Mistral-7B}   & PI                                       & 2.20                                     & 2.16                                     & -                                        & -                                        & -                                         \\
                              & InfLM                                    & 2.43                                     & 2.40                                     & 2.41                                     & 2.42                                     & 2.41                                      \\
                              & YARN                                     & 2.20                                     & 2.16                                     & 2.17                                     & 2.16                                     & 2.16                                      \\
                              & {\cellcolor[rgb]{0.863,0.863,0.863}}Tens & {\cellcolor[rgb]{0.863,0.863,0.863}}2.18 & {\cellcolor[rgb]{0.863,0.863,0.863}}2.16 & {\cellcolor[rgb]{0.863,0.863,0.863}}2.15 & {\cellcolor[rgb]{0.863,0.863,0.863}}2.15 & {\cellcolor[rgb]{0.863,0.863,0.863}}2.16  \\
                              & Time                                     & 0.40$\times$                             & 0.23$\times$                             & 0.13$\times$                             & 0.09$\times$                             & 0.07$\times$                              \\ 
\midrule
\multirow{5}{*}{Llama-8B}     & PI                                       & 2.22                                     & 2.20                                     & -                                        & -                                        & -                                         \\
                              & InfLM                                    & 2.45                                     & 2.34                                     & 2.36                                     & 2.41                                     & 2.39                                      \\
                              & YARN                                     & 2.22                                     & 2.20                                     & 2.27                                     & 2.29                                     & 2.28                                      \\
                              & {\cellcolor[rgb]{0.863,0.863,0.863}}Tens & {\cellcolor[rgb]{0.863,0.863,0.863}}2.30 & {\cellcolor[rgb]{0.863,0.863,0.863}}2.29 & {\cellcolor[rgb]{0.863,0.863,0.863}}2.21 & {\cellcolor[rgb]{0.863,0.863,0.863}}2.19 & {\cellcolor[rgb]{0.863,0.863,0.863}}2.16  \\
                              & Time                                     & 0.72$\times$                             & 0.25$\times$                             & 0.15$\times$                             & 0.11$\times$                             & 0.09$\times$                              \\
\bottomrule
\end{tabular}
}
\caption{Extrapolation perplexity and efficiency with sequence length ranging from 16k to 128k. "-" denotes perplexity $> 10$. Time usage is compared between tensorized and full attention.
}

\label{tab::extrapo}
\end{table}

\vspace{-3pt}
\section{Conclusion}
\vspace{-2pt}
In this paper, we explore attention tensorization by replacing vector/matrix operations with corresponding tensor operations. 
From an efficiency perspective, decomposing long sequences into individual dimensions allows the limited context length budget to be utilized more effectively for attention calculation along each dimension.
From a performance perspective, attention tensorization naturally encodes multi-hop hierarchical token interactions, making attention more efficiently approximated compared to its representation in vector space. 
Extensive results on datasets from various domains demonstrate that tensorization can extend the training context length, enhancing both efficiency and length extrapolation performance.

\section{Acknowledgments}
This work is supported by the U.S. Department of Energy under award DE-FOA-0003264 and the Army Research Office under grant W911NF-23-1-0088.
%%%%%%%%%%%%%%%%%%%%%%%%%%%%%%%%%%%%%%%%%%%%%%%%%%%%%%%%%%%%
\newpage

\bibliography{main}
\newpage
\appendix

\section{Attention Patterns in Transformers}
\begin{figure}[H]
  \centering \includegraphics[width=0.95\linewidth]{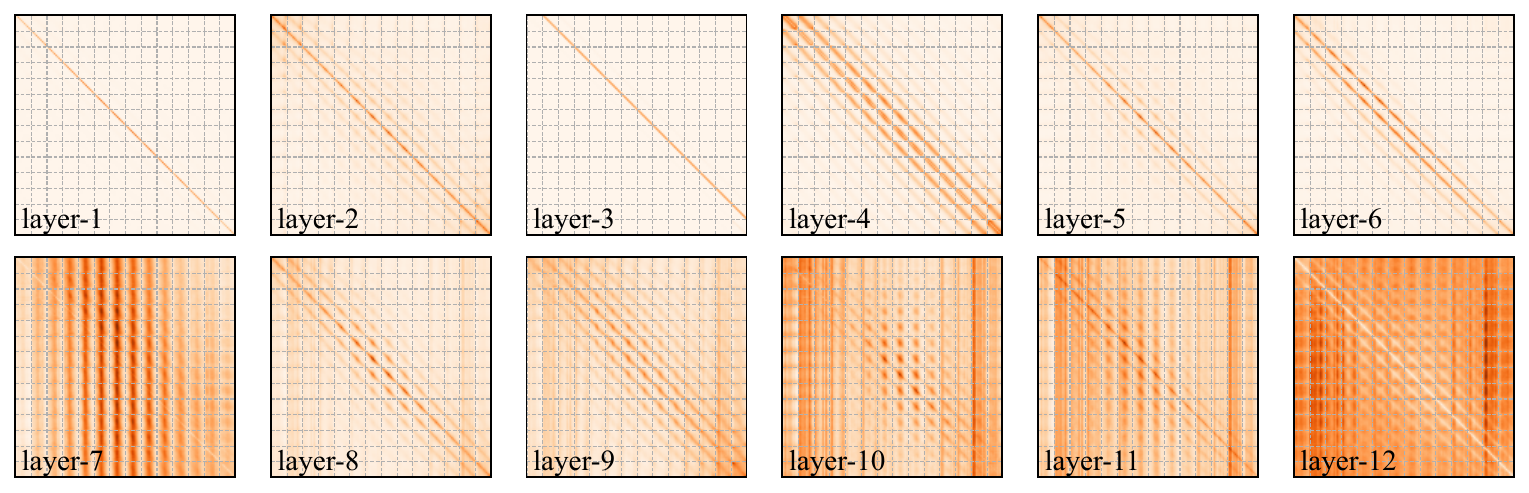}
  \caption{Averaged attention patterns at different layers with the 6th head.
  }
\label{sfig::attpattern_layer}
\end{figure}

\begin{figure}[H]
  \centering \includegraphics[width=0.95\linewidth]{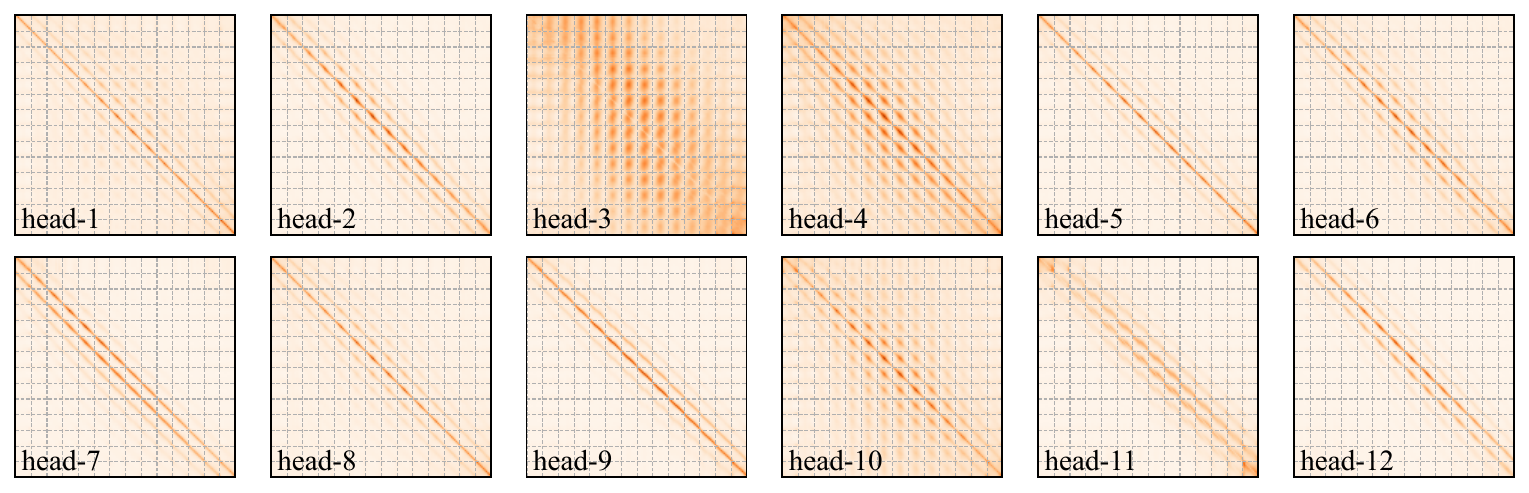}
  \caption{Averaged attention patterns at different heads at the 5th layer.
  }
\label{sfig::attpattern_head}
\end{figure}

\begin{figure}[H]
  \centering \includegraphics[width=0.95\linewidth]{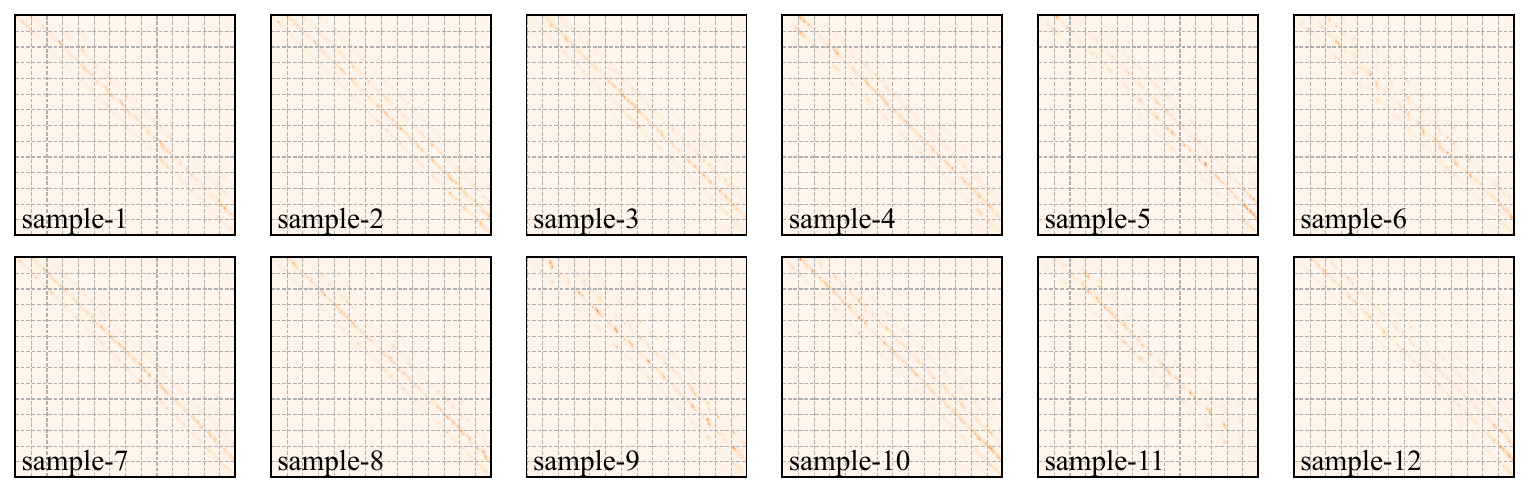}
  \caption{Attention patterns of different samples at the 3rd layer and 0th head.
   }
\label{sfig::attpattern_sample}
\end{figure}

In this section, we show the hierarchical and low-rank attention structure using empirical results.
We show the attention patterns of the IMDB validation set using a trained RoBERTa model. 
As can be seen, shallow layers show more diagonal patterns while deeper layers have more block and column patterns (Figure \ref{sfig::attpattern_layer}).
Different heads in the same layer have different patterns that attend to different aspects of data (Figure \ref{sfig::attpattern_head}).
Different samples will induce slightly different patterns to attend to sample-specific details (Figure \ref{sfig::attpattern_sample}).

\section{More Spectrum Analysis}
\subsection{Spectrum on ViT attention}
\label{asec::nlp_spec}
\begin{figure}[H]
  \centering \includegraphics[width=0.99\linewidth]{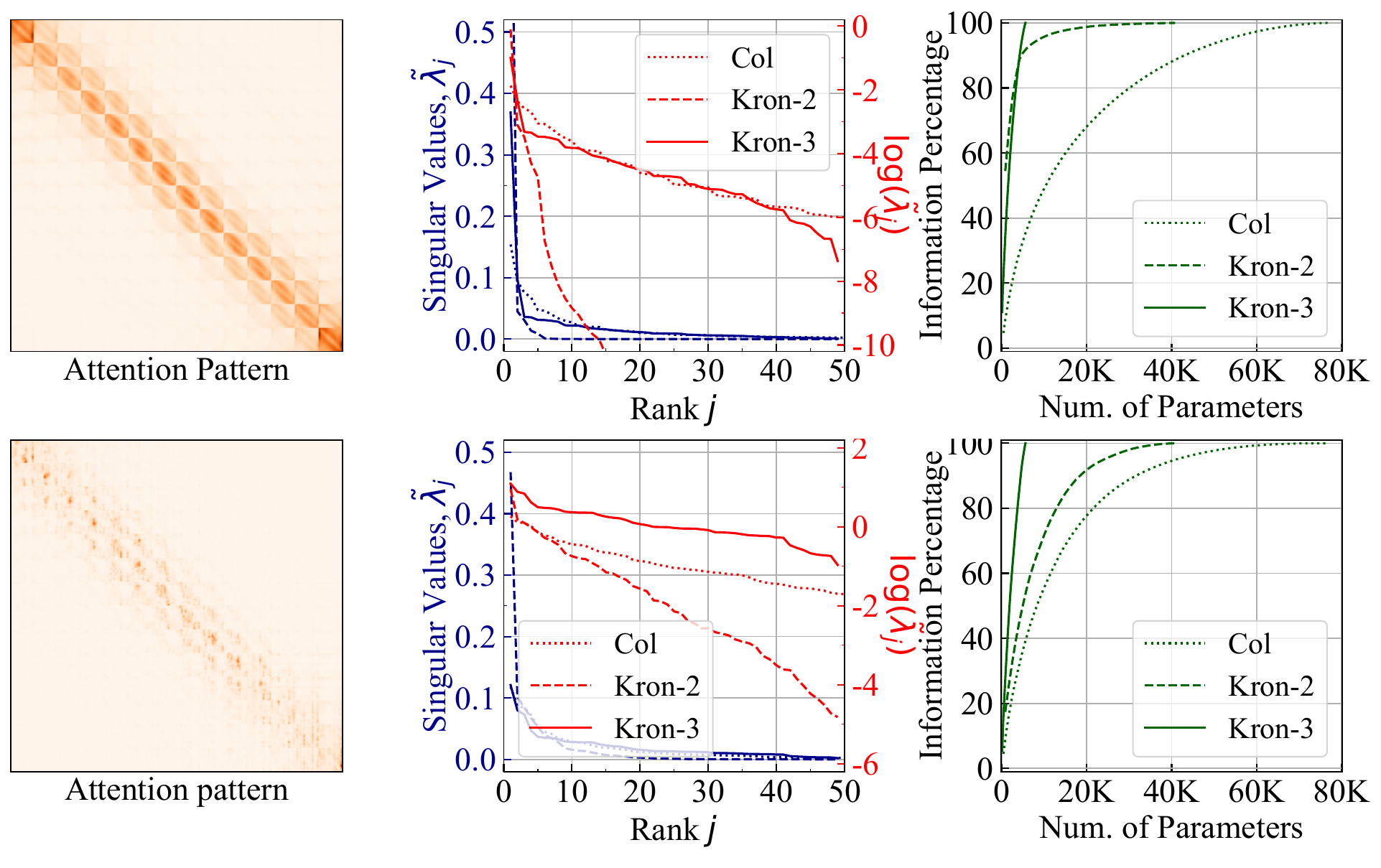}
  \caption{Singular value spectrum of averaged and single attention patterns with NLP dataset.
  }
\label{sfig::nlp}
\end{figure}

We perform the low-rank approximation analysis on ViT-Base model with Imagenet-1K dataset, similar to the image model analysis in Section \ref{sec::spectrum}. 
As shown in Figure \ref{sfig::nlp}, compared to vector space, we can obtain approximations with lower rank and fewer parameters in the proposed tensor space.

\subsection{Theorem \ref{thm::spectrum}}
\label{asec::theo}
The theorem shows that the required rank can decay exponentially with $m\geq1$, and as we increase the Kronecker decomposition order $m$, the exponential decay will be faster.
Since in practice, $m$ is usually very small (usually 2 - 5) while $n$ can be over thousands, the main factor in the rank complexity is from term $\text{log}_{2m}n$, indicating the exponential contribution of $m$.

\xhdr{Proof} We follow \cite{wang2020linformer} to prove the theorem using the distributional Johnson-Lindenstrauss (JL) lemma \cite{johnson1984extensions}.
Defining the low-rank approximation matrix $\tilde{\mat{A}}=\mat{A}\mat{T}^\intercal\mat{T}$, where $\mat{T}\in\mathbb{R}^{k\times n}$ and $k<n$. Then we get $\text{rank}(\tilde{\mat{A}})\leq \text{rank}(\mat{T})\leq k$.

Assuming tensor-$m$ space is constructed from block sizes $\{ n_i\}_{i=1}^{m}$ with $\prod_{i=1}^{m}n_i=n$.
Then for $i$-th row vector $\ve{t}^{(i)}\in\mathbb{R}^{n}$ of $\mat{T}$ and $\ve{y}$, the corresponding representations in tensor-$m$ space is  $\ten{T}^{(i)}\in\mathbb{R}^{n_1\times ...n_m}$ and $\ten{Y}\in\mathbb{R}^{n_1\times ...n_m}$.
We additionally assume  $\ten{T}^{(i)}$ has rank-$r$: $\ten{T}^{(i)}=\sum_{j=1}^{r}\otimes_{h=1}^{m}\ve{t}^{(i)}_{h,j}$.

Next, we consider the transformation induced by $\mat{T}$ in the row-by-row form as 
\begin{equation}
    \mat{T}\ve{y}=\begin{bmatrix}
                    \left<\ve{t}^{(1)}, \ve{y}\right> \\
                    ...\\
                    \left<\ve{t}^{(k)}, \ve{y}\right>
                    \end{bmatrix},
\end{equation}
where each row can be calculated using mixed Kronecker-matrix(vector) product property as
\begin{equation}
    \left<\ve{t}^{(i)}, \ve{y}\right> = \left<\ten{T}^{(i)}, \ten{Y}\right>
\end{equation}
with the general inner product defined as $\left<\ten{A},\ten{B}\right>=\sum_{i_1,...,i_m}\ten{A}_{i_1,...,i_m}\ten{B}_{i_1,...,i_m}$.

JL properties of this transformation $\mat{T}\ve{y}$ can be then characterized by the following lemma:
\begin{lemma} (Theorem 2 in \cite{rakhshan2020tensorized})
Let $\mat{T}$ be $k\times n$ matrix with i.i.d  entries from $N(0, r^{-\frac{1}{m}})$, for any $\epsilon>0, \delta>0$, if $k>\epsilon^{-2}3^{m-1}(1+2/R)\text{log}_{2m}(N/\delta)$, we have 
\begin{equation}
Pr(| \lVert\mat{T}\ve{y}\rVert-\lVert\ve{y}\rVert | > \epsilon\lVert\ve{y}\rVert)< \delta   
\end{equation}
\begin{equation}
Pr(| \lVert\ve{x}^\intercal\mat{T}^\intercal \mat{T}\ve{y}\rVert-\lVert \ve{x}^\intercal\ve{y} \rVert | > \epsilon\lVert \ve{x}^\intercal\ve{y} \rVert)< 2\delta   .
\end{equation}
\end{lemma}

Therefore, we have
\begin{equation}
\begin{split}
&Pr\left(\lVert\tilde{\mat{A}}\ve{y} -\mat{A}\ve{y} \rVert <\epsilon\lVert\mat{A}\ve{y} \rVert\right)\\
&=Pr\left(\lVert\mat{A}\mat{T}^\intercal\mat{T}\ve{y} -\mat{A}\ve{y} \rVert <\epsilon\lVert\mat{A}\ve{y} \rVert\right) \\
&\geq1-\\&\sum_{i=1}^{n}Pr\left(| \lVert\ve{a}^{(i)\intercal}\mat{T}^\intercal \mat{T}\ve{y}\rVert-\lVert \ve{a}^{(i)\intercal}\ve{y} \rVert | > \epsilon\lVert \ve{a}^{(i)\intercal}\ve{y} \rVert\right) \\
&\geq 1-2n\delta.
\end{split}
\end{equation}
Let $\delta=\frac{1}{n}$ and hence $\text{rank}(\tilde{\mat{A}})=k=\mathcal{O}(3^{m}\text{log}_{2m}n)$, the theorem follows.

\section{More Discussions of Attention Tensorization}
\subsection{Tensorization with Reshaping}
\label{asec::reshape}
\begin{figure}[H]
  \centering \includegraphics[width=0.95\linewidth]{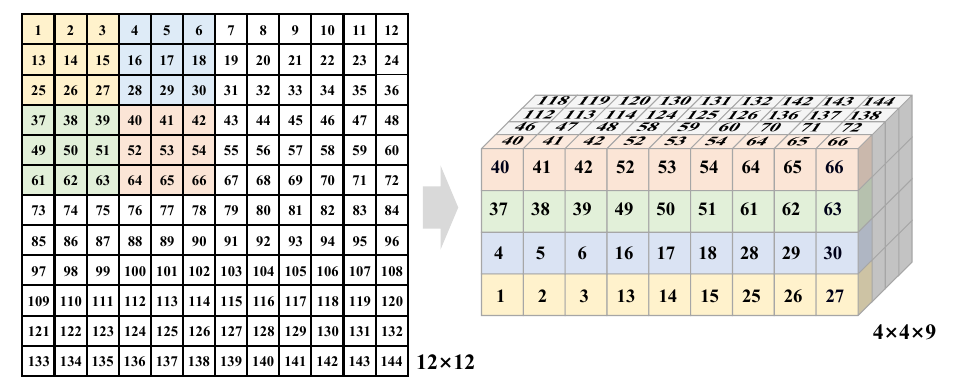}
  \caption{Tensorization from a $12\times 12$ matrix to $4\times 4\times 9$ tensor in tensor space (constructed from order-3 Kronecker decomposition of the matrix).
  }
\label{sfig::tensorization}
\end{figure}

As mentioned, to map an $n$-by-$n$ attention matrix $\mat{A}$ into its representation $\ten{A}$ in tensor space through tensorization, we simply need to reshape the attention matrix.
We show a tensorization example in Figure \ref{sfig::tensorization} to better understand the construction of tensor space.
The yellow row dimension in $\ten{A}$ is from reshaping the yellow square block of $\mat{A}$.
It can be seen that row, column, and height dimensions in $\ten{A}$ represent low-, middle-, and high-level attention in $\mat{A}$ respectively.

\subsection{Time Complexity}
\label{asec::time_complexity}
We compare the theoretical time complexity of the proposed tensorization mechanism at the three attention steps with other popular attention approximation methods as in Table \ref{stab::time_complexity}.
\begin{table}[H]

\centering
\resizebox{1.0\columnwidth}{!}{
\begin{tabular}{llll} 
\toprule
Operations      & $\ve{q}\ve{k}^\intercal$ & softmax                                          & $\mat{A}\ve{v}$     \\ 
\midrule
Full          & $\mathcal{O}(n^2d)$                                      & $\mathcal{O}(n^2)$             & $\mathcal{O}(n^2d)$              \\
Sparse        & $\mathcal{O}(pn^2d)$                                     & $\mathcal{O}(pn^2)$            & $\mathcal{O}(pn^2d)$              \\
Linformer     & $\mathcal{O}(nkd) $                                         & $\mathcal{O}(nk)$                 & $\mathcal{O}(nkd)$                   \\
Performer     & $\mathcal{O}(nkd)$                                          & $\mathcal{O}(nk)$                 & $\mathcal{O}(nkd)$                   \\
Reformer      & $\mathcal{O}(n_r(4n/n_c)^2d)$                          &$ \mathcal{O}(n_r(4n/n_c)^2)$ & $\mathcal{O}(n_rn(4n/n_c)^2d)$  \\
Nystromformer & $\mathcal{O}(nkd+k^2d) $                                 & $\mathcal{O}(nk+k^2)   $       & $\mathcal{O}(nk^2+2nkd+k^3)$   \\
Tensorization       & $\mathcal{O}(\sum_{i=1}^{m} p_in_i^2d)  $                                            &    $\mathcal{O}(\sum_{i=1}^{m} p_in_i^2)$                                              &               $\mathcal{O}( pn\sum_{i=1}^{m}n_i d)$                                      \\ %%$
\bottomrule
\end{tabular}
}
\caption{Time complexity of different efficient attention mechanisms, with length dimension $n, k$, feature dimension $d$, sparsity $p$. }
\vspace{-5pt}
\label{stab::time_complexity}
\end{table}

\section{Experiments}
We present additional experimental results of tensorized attention across various datasets, including both fine-tuning and training from scratch settings.
\label{asec::experiment}
\subsection{Language Modeling with RoBERTa}
\label{asec::roberta_GLUE}
Given pretrained full-attention model, tensorized attention can be directly adopted as an efficient method for finetuning, with improved training and inference time.
We evaluate the finetuning performance of tensorized attention ith RoBERTa backbones on GLUE \cite{wang2018glue} benchmark. 
We follow the finetuning setting of RoBERTa on GLUE development as \cite{liu2019roberta}, and initialize the model with pretrianed RoBERTa on every task (without using MNLI results to initialize MRPC, RTE, and STS-B).
Full parameter and sparse fituning with different attention masks (90\% masked) are used as benchmarks.
We use "Tens-$n$" to denote the $n$-th order tensorization.
Median results for each task over five random seeds are reported in Table \ref{stab::glue_res}.
The optimal tensorization order is different among tasks and determined by the input data type. On average, RoBERTa model adapted with tensorized attention shows on-par and better results compared to full-attention baselines while only a small fraction of the attention calculations is needed. 
\begin{table}[H]
\centering
\resizebox{0.99\columnwidth}{!}{
\begin{tabular}{cl|ccccccccc} 
\toprule
\multicolumn{2}{c|}{Dataset}     & MNLI          & QQP           & QNLI          & SST-2         & CoLA          & STS-B        & MRPC          & RTE           & Average         \\ 
\hline
\multicolumn{2}{c|}{Full}        & \textbf{87.6} & \uline{91.6}  & 92.0          & \uline{94.6}  & \uline{63.8}  & \uline{91.2} & \textbf{90.0} & 80.2          & \uline{86.38}   \\ 
\hline
\multirow{3}{*}{Sparse} & Window & 82.4          & 88.2          & 87.6          & 92.5          & 58.6          & 86.2         & 83.6          & 76.5          & 81.95           \\
                        & Rand   & 79.8          & 85.8          & 84.9          & 91.5          & 54.3          & 83.3         & 82.8          & 73.4          & 79.48           \\
                        & Top-k  & 80.3          & 88.0          & 88.2          & 92.4          & 59.8          & 86.5         & 86.2          & 79.2          & 82.58           \\ 
\hline
\multicolumn{2}{c|}{Tens-1}      & 82.5          & 89.5          & 89.5          & 93.1          & 59.5          & 87.8         & 86.0          & 78.7          & 83.33           \\
\multicolumn{2}{c|}{Tens-2}      & \uline{87.2}  & 91.1          & \textbf{92.4} & \textbf{94.8} & \textbf{64.1} & \uline{91.2} & \uline{89.5}  & \textbf{86.4} & \textbf{87.09}  \\
\multicolumn{2}{c|}{Tens-3}      & 86.8          & \textbf{91.8} & \uline{92.2}  & 94.5          & 62.6          & 90.9         & 86.2          & \uline{84.5}  & 86.19           \\
\bottomrule
\end{tabular}
}
\caption{Comparison of fine-tuning RoBERTa model using full, sparse, and tensorization on the GLUE development set (\%). 
Underline values are second best, and bold values are the best.}
\label{stab::glue_res}
\end{table}

\subsection{Long Time Series Modeling}

We evaluate the performance of attention tensorization in the time series domain under training from scratch setting.
We choose SOTA time series transformers PatchTST \cite{nie2022time} as the backbone, and replace the full attention in the attention layer with tensorization.
The resulting Tens time series model is compared with SOTA baselines 1-NN (ED)\cite{ruiz2021great}, ResNet18\cite{ismail2019deep}, RandomShapelet\cite{karlsson2016generalized}, DLinear \cite{zeng2023transformers}, Temporal Convolutional Network \cite{bai2018empirical}, and PatchTST.

From the UEA Time Series Classification Archive, we choose the following fine-resolution time series datasets BinaryHeartbeat, EigenWorms, FaultDetectionA, CatsDogs, each has lengths greater than 10,000.
For each dataset, we uses $80\%$ of data as training data, $10\%$ of data as validation data, and the rest $10\%$ of data as testing data.
The final test accuracy is obtained using the checkpoint of the lowest validation loss, with the early-stop patience epochs 20.

\begin{table}[h]

\resizebox{0.99\columnwidth}{!}{
\centering
\begin{tabular}{l cccc} \toprule
 \cmidrule{2-5}
\multicolumn{1}{l}{Model} & \multicolumn{1}{l}{BinaryHeartbeat} & \multicolumn{1}{l}{EigenWorms} & \multicolumn{1}{l}{FaultDetectionA} & \multicolumn{1}{l}{CatsDogs} \\  \cmidrule{1-5}
1-NN (ED)      & 0.585                                & 0.500                           & 0.460                           & 0.420                                 \\
ResNet18       & 0.630                                & 0.420                           & 0.990                           & 0.570                                 \\
RandomShapelet & 0.585                                & 0.692                           & N/A                           & 0.606                        \\
DLinear        & 0.658                                & 0.423                           & 0.532                           & 0.516                                 \\
TCN            & \textbf{0.707}                                & \textbf{0.769}                  & 0.986                           & 0.575                                 \\
PatchTST       & \textbf{0.707}                       & 0.692                           &           0.991        & 0.810                        \\  
Tens & \textbf{0.707}                      & \textbf{0.769}                           & \textbf{0.996}             & \textbf{0.844}                                 \\    \bottomrule
\end{tabular}
}
\caption{Time series classification performance comparison.}
\label{stab::timeseries}
\end{table}

\begin{figure}[ptb]
\centering
\begin{subfigure}[b]{.98\linewidth}
\input{fig_eff1}
\label{fig:tokenlength_amazon}
\end{subfigure}

\caption{Comparison of running timewith different input lengths.}
\label{sfig::efficiency}
% \vspace{-4mm}
\end{figure}
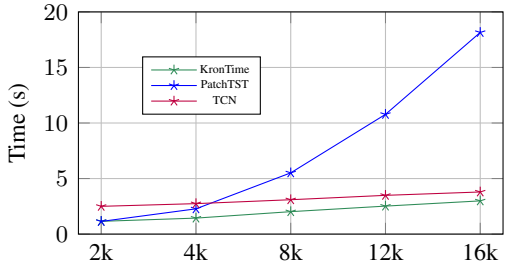

The classification result is shown in Table \ref{stab::timeseries} with efficiency shown in Figure \ref{sfig::efficiency}. Tensorization achieves the same or superior classification accuracy compared to SOTA models. 
The results demonstrate that tensorization achieves approximately $0.3\times$ running time compared to PatchTST at a length of 16k, and is comparable to the conventional convolution-based TCN. This advantage of improved running time from attention decomposition becomes more significant as the input length increases.

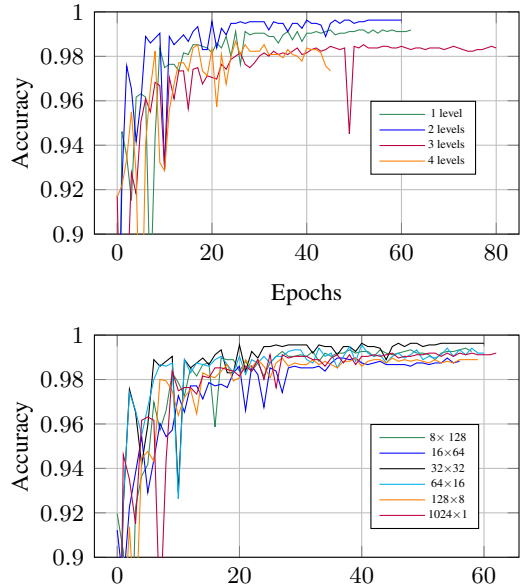
\begin{figure}
\centering

\begin{subfigure}[b]{.98\linewidth}
\input{fig_abl_level}

\end{subfigure}

\begin{subfigure}[b]{.98\linewidth}
\input{fig_abl_size}
\label{fig:tokenlength_ildc}
\end{subfigure}
\caption{The validation accuracy with different tensorization strategies (upper: number of levels decomposed; lower: different tensorization with 2 levels) during the training phase.}
\label{sfig:timeseries_abla}
\end{figure}

We next demonstrate the influence of the tensorization strategies by comparing the training curves.
We perform such ablation studies on FaultDetectionA dataset with 1024 input length after tokenized.
We change the total number of orders (levels) of tensorization and the size of each level while keeping other model and training hyperparameters unchanged for fair comparisons.
As shown in Figure \ref{sfig:timeseries_abla}(a), training with 2-level tensorization ($16\times 16$) converges to the higher validation accuracy with faster speed, compared to no tensorization (1-level tensorization), 3-level tensorization ($16\times 16\times 4$), and 4-level tensorization ($16\times 4\times4\times 4$).
We then compare different tensorization strategies with 2 levels, and results in Figure \ref{sfig:timeseries_abla}(b) show that tensorizing the $1024$ sequence into $32\times 32$ achieves superior results compared to other tensorizations scheme for this dataset.

\end{document}

%% file: fig_eff1.tex
\begin{tikzpicture}
\centering
\tikzstyle{every node}=[font=\small]
%https://tex.stackexchange.com/questions/509171/how-to-highlight-a-point-in-line-graph-by-drawing-a-circle-around-it
\begin{axis}[
    width  = 0.95\linewidth,
    height = 0.6\linewidth,
    ymin = 0, ymax = 20,
    symbolic x coords = {2k, 4k, 8k, 12k, 16k},
    ytick = {},
    ylabel = {Time (s)},
    ymajorgrids = true,
    xmajorgrids = true,
    y label style={at={(0.08,0.5)}},
    enlarge x limits={abs=0.3cm},
     legend style={
        at={(0.15,0.8)},
        anchor=north west,
        nodes={scale=0.5, transform shape},
        }
]
\addplot[color = bigbird_color, mark = star ]
    coordinates {
    (2k, 1.15)
    (4k, 1.44)
    (8k, 2.02)
    (12k, 2.52)
    (16k, 2.99)
    };
\addlegendentry{KronTime}

\addplot[color = blue, mark = star]
    coordinates {
    (2k, 1.14)
    (4k, 2.275)
    (8k, 5.513)
    (12k, 10.778)
    (16k, 18.151)
    };
\addlegendentry{PatchTST}

\addplot[color = purple, mark = star]
    coordinates {
    (2k, 2.4998)
    (4k, 2.7426)
    (8k, 3.0947)
    (12k, 3.4885)
    (16k, 3.7916)
    };
\addlegendentry{TCN}
\end{axis}
\end{tikzpicture}

%% file: fig_abl_level.tex
\begin{tikzpicture}
\centering
\tikzstyle{every node}=[font=\small]
%https://tex.stackexchange.com/questions/509171/how-to-highlight-a-point-in-line-graph-by-drawing-a-circle-around-it
\begin{axis}[
    width  = 0.95\linewidth,
    height = 0.6\linewidth,
    ymin = 0.9, ymax = 1,
    % x coords = {0, 20, 40, 60, 80, 100},
    xtick = {0, 20, 40, 60, 80, 100},
    ytick = {},
    xlabel = {Epochs},
    ylabel = {Accuracy},
    ymajorgrids = true,
    xmajorgrids = true,
    y label style={at={(0.05,0.5)}},
    enlarge x limits={abs=0.3cm},
     legend style={
        at={(0.65,0.6)},
        anchor=north west,
        nodes={scale=0.5, transform shape},
        }
]

\addplot[color = bigbird_color, mark = circle ]
    coordinates {
    (0, 0.7544247508049011)
    (1, 0.9461652040481567)
    (2, 0.9351032376289368)
    (3, 0.9151917099952698)
    (4, 0.9616519212722778)
    (5, 0.9631268382072449)
    (6, 0.9616519212722778)
    (7, 0.8709439635276794)
    (8, 0.9439527988433838)
    (9, 0.983775794506073)
    (10, 0.974926233291626)
    (11, 0.976401150226593)
    (12, 0.976401150226593)
    (13, 0.9734513163566589)
    (14, 0.9815633893013)
    (15, 0.9808259606361389)
    (16, 0.98525071144104)
    (17, 0.98525071144104)
    (18, 0.9845132827758789)
    (19, 0.983775794506073)
    (20, 0.9815633893013)
    (21, 0.985988199710846)
    (22, 0.983775794506073)
    (23, 0.98893803358078)
    (24, 0.980088472366333)
    (25, 0.991150438785553)
    (26, 0.976401150226593)
    (27, 0.991150438785553)
    (28, 0.9896755218505859)
    (29, 0.9904129505157471)
    (30, 0.9904129505157471)
    (31, 0.9904129505157471)
    (32, 0.98893803358078)
    (33, 0.98893803358078)
    (34, 0.985988199710846)
    (35, 0.9896755218505859)
    (36, 0.991150438785553)
    (37, 0.987463116645813)
    (38, 0.991150438785553)
    (39, 0.9867256879806519)
    (40, 0.9904129505157471)
    (41, 0.9904129505157471)
    (42, 0.991150438785553)
    (43, 0.9904129505157471)
    (44, 0.991150438785553)
    (45, 0.9918879270553589)
    (46, 0.9896755218505859)
    (47, 0.9904129505157471)
    (48, 0.9904129505157471)
    (49, 0.991150438785553)
    (50, 0.991150438785553)
    (51, 0.9918879270553589)
    (52, 0.9918879270553589)
    (53, 0.9904129505157471)
    (54, 0.9918879270553589)
    (55, 0.9904129505157471)
    (56, 0.9918879270553589)
    (57, 0.991150438785553)
    (58, 0.9918879270553589)
    (59, 0.991150438785553)
    (60, 0.991150438785553)
    (61, 0.991150438785553)
    (62, 0.9918879270553589)
    };
\addlegendentry{1 level}

\addplot[color = blue, mark = circle]
    coordinates {
(0, 0.8230088353157043)
(1, 0.9233038425445557)
(2, 0.9756637215614319)
(3, 0.9653392434120178)
(4, 0.9417403936386108)
(5, 0.9594395160675049)
(6, 0.98893803358078)
(7, 0.985988199710846)
(8, 0.9882006049156189)
(9, 0.9904129505157471)
(10, 0.9292035102844238)
(11, 0.98893803358078)
(12, 0.98525071144104)
(13, 0.987463116645813)
(14, 0.9896755218505859)
(15, 0.9867256879806519)
(16, 0.991150438785553)
(17, 0.9933628439903259)
(18, 0.9830383658409119)
(19, 0.9830383658409119)
(20, 0.9955751895904541)
(21, 0.983775794506073)
(22, 0.99262535572052)
(23, 0.9896755218505859)
(24, 0.994837760925293)
(25, 0.994837760925293)
(26, 0.9955751895904541)
(27, 0.994837760925293)
(28, 0.9955751895904541)
(29, 0.9955751895904541)
(30, 0.9955751895904541)
(31, 0.9955751895904541)
(32, 0.9918879270553589)
(33, 0.9955751895904541)
(34, 0.994837760925293)
(35, 0.994837760925293)
(36, 0.991150438785553)
(37, 0.994837760925293)
(38, 0.994837760925293)
(39, 0.99262535572052)
(40, 0.99631267786026)
(41, 0.994837760925293)
(42, 0.994837760925293)
(43, 0.9941002726554871)
(44, 0.98893803358078)
(45, 0.994837760925293)
(46, 0.99631267786026)
(47, 0.994837760925293)
(48, 0.99631267786026)
(49, 0.99631267786026)
(50, 0.994837760925293)
(51, 0.9955751895904541)
(52, 0.9955751895904541)
(53, 0.99631267786026)
(54, 0.99631267786026)
(55, 0.99631267786026)
(56, 0.99631267786026)
(57, 0.99631267786026)
(58, 0.99631267786026)
(59, 0.99631267786026)
(60, 0.99631267786026)
    };
    
\addlegendentry{2 levels}

\addplot[color = purple, mark = circle]
    coordinates {
(0, 0.9174041152000427)
(1, 0.8215339183807373)
(2, 0.8672566413879395)
(3, 0.9277285933494568)
(4, 0.9181416034698486)
(5, 0.9505899548530579)
(6, 0.9609144330024719)
(7, 0.9550147652626038)
(8, 0.9682890772819519)
(9, 0.9668141603469849)
(10, 0.9299409985542297)
(11, 0.9705014824867249)
(12, 0.9609144330024719)
(13, 0.9734513163566589)
(14, 0.9734513163566589)
(15, 0.9564896821975708)
(16, 0.9741888046264648)
(17, 0.974926233291626)
(18, 0.967551589012146)
(19, 0.971238911151886)
(20, 0.9705014824867249)
(21, 0.969763994216919)
(22, 0.976401150226593)
(23, 0.9741888046264648)
(24, 0.978613555431366)
(25, 0.980088472366333)
(26, 0.9815633893013)
(27, 0.976401150226593)
(28, 0.974926233291626)
(29, 0.9771386384963989)
(30, 0.980088472366333)
(31, 0.9815633893013)
(32, 0.980088472366333)
(33, 0.982300877571106)
(34, 0.9808259606361389)
(35, 0.9815633893013)
(36, 0.9808259606361389)
(37, 0.982300877571106)
(38, 0.983775794506073)
(39, 0.9808259606361389)
(40, 0.9815633893013)
(41, 0.9830383658409119)
(42, 0.983775794506073)
(43, 0.982300877571106)
(44, 0.9830383658409119)
(45, 0.9845132827758789)
(46, 0.983775794506073)
(47, 0.9830383658409119)
(48, 0.983775794506073)
(49, 0.945132827758789)
(50, 0.9845132827758789)
(51, 0.983775794506073)
(52, 0.98525071144104)
(53, 0.983775794506073)
(54, 0.983775794506073)
(55, 0.983775794506073)
(56, 0.9845132827758789)
(57, 0.9845132827758789)
(58, 0.983775794506073)
(59, 0.983775794506073)
(60, 0.9845132827758789)
(61, 0.983775794506073)
(62, 0.9830383658409119)
(63, 0.983775794506073)
(64, 0.983775794506073)
(65, 0.9830383658409119)
(66, 0.983775794506073)
(67, 0.983775794506073)
(68, 0.983775794506073)
(69, 0.9830383658409119)
(70, 0.983775794506073)
(71, 0.9845132827758789)
(72, 0.983775794506073)
(73, 0.9830383658409119)
(74, 0.982300877571106)
(75, 0.982300877571106)
(76, 0.9830383658409119)
(77, 0.9830383658409119)
(78, 0.983775794506073)
(79, 0.9845132827758789)
(80, 0.983775794506073)
    };
\addlegendentry{3 levels}

\addplot[color = orange, mark = circle]
    coordinates {
(0, 0.9166666865348816)
(1, 0.9218289256095886)
(2, 0.9336283206939697)
(3, 0.9550147652626038)
(4, 0.9269911646842957)
(5, 0.8340708017349243)
(6, 0.9410029649734497)
(7, 0.9653392434120178)
(8, 0.982300877571106)
(9, 0.9321534037590027)
(10, 0.9284660816192627)
(11, 0.9557521939277649)
(12, 0.967551589012146)
(13, 0.9771386384963989)
(14, 0.9771386384963989)
(15, 0.9734513163566589)
(16, 0.983775794506073)
(17, 0.98525071144104)
(18, 0.9734513163566589)
(19, 0.971238911151886)
(20, 0.9830383658409119)
(21, 0.9572271108627319)
(22, 0.9808259606361389)
(23, 0.967551589012146)
(24, 0.980088472366333)
(25, 0.9867256879806519)
(26, 0.9808259606361389)
(27, 0.98525071144104)
(28, 0.982300877571106)
(29, 0.982300877571106)
(30, 0.982300877571106)
(31, 0.98525071144104)
(32, 0.9808259606361389)
(33, 0.9808259606361389)
(34, 0.9808259606361389)
(35, 0.9793510437011719)
(36, 0.9830383658409119)
(37, 0.9808259606361389)
(38, 0.980088472366333)
(39, 0.9778761267662048)
(40, 0.983775794506073)
(41, 0.9830383658409119)
(42, 0.9830383658409119)
(43, 0.9815633893013)
(44, 0.9756637215614319)
(45, 0.9734513163566589)
    };
\addlegendentry{4 levels}

\end{axis}
\end{tikzpicture}

%% file: fig_abl_size.tex
\begin{tikzpicture}
\centering
\tikzstyle{every node}=[font=\small]
%https://tex.stackexchange.com/questions/509171/how-to-highlight-a-point-in-line-graph-by-drawing-a-circle-around-it
\begin{axis}[
    width  = 0.95\linewidth,
    height = 0.6\linewidth,
    ymin = 0.9, ymax = 1,
    % symbolic x coords = {20, 40, 60, 80, 100},
    % x coords = {0, 20, 40, 60, 80, 100},
    xtick = {0, 20, 40, 60, 80, 100},
    ytick = {},
    ylabel = {Accuracy},
    ymajorgrids = true,
    xmajorgrids = true,
    y label style={at={(0.05,0.5)}},
    enlarge x limits={abs=0.3cm},
     legend style={
        at={(0.65,0.6)},
        anchor=north west,
        nodes={scale=0.5, transform shape},
        }
]
\addplot[color = bigbird_color, mark = circle ]
    coordinates {
(0, 0.9196165204048157)
(1, 0.9026548862457275)
(2, 0.8222714066505432)
(3, 0.9225663542747498)
(4, 0.9365781545639038)
(5, 0.9417403936386108)
(6, 0.969763994216919)
(7, 0.9572271108627319)
(8, 0.966076672077179)
(9, 0.9830383658409119)
(10, 0.978613555431366)
(11, 0.9727138876914978)
(12, 0.98525071144104)
(13, 0.9815633893013)
(14, 0.9867256879806519)
(15, 0.987463116645813)
(16, 0.9587020874023438)
(17, 0.9896755218505859)
(18, 0.98893803358078)
(19, 0.98893803358078)
(20, 0.983775794506073)
(21, 0.9882006049156189)
(22, 0.983775794506073)
(23, 0.987463116645813)
(24, 0.9815633893013)
(25, 0.987463116645813)
(26, 0.9904129505157471)
(27, 0.99262535572052)
(28, 0.9904129505157471)
(29, 0.9904129505157471)
(30, 0.991150438785553)
(31, 0.9904129505157471)
(32, 0.99262535572052)
(33, 0.9882006049156189)
(34, 0.9918879270553589)
(35, 0.9904129505157471)
(36, 0.9918879270553589)
(37, 0.9918879270553589)
(38, 0.9933628439903259)
(39, 0.9918879270553589)
(40, 0.99262535572052)
(41, 0.99262535572052)
(42, 0.9904129505157471)
(43, 0.9933628439903259)
(44, 0.9918879270553589)
(45, 0.991150438785553)
(46, 0.9918879270553589)
(47, 0.99262535572052)
(48, 0.9933628439903259)
(49, 0.9918879270553589)
(50, 0.9918879270553589)
(51, 0.991150438785553)
(52, 0.98893803358078)
(53, 0.9918879270553589)
(54, 0.99262535572052)
(55, 0.99262535572052)
(56, 0.9933628439903259)
(57, 0.9941002726554871)
(58, 0.99262535572052)
    };
\addlegendentry{8$\times$ 128}

\addplot[color = blue, mark = circle]
    coordinates {
(0, 0.9122418761253357)
(1, 0.8871681094169617)
(2, 0.9225663542747498)
(3, 0.9380530714988708)
(4, 0.9542772769927979)
(5, 0.9292035102844238)
(6, 0.9439527988433838)
(7, 0.9601770043373108)
(8, 0.9542772769927979)
(9, 0.9572271108627319)
(10, 0.9727138876914978)
(11, 0.9653392434120178)
(12, 0.9771386384963989)
(13, 0.9771386384963989)
(14, 0.971238911151886)
(15, 0.978613555431366)
(16, 0.9771386384963989)
(17, 0.9778761267662048)
(18, 0.976401150226593)
(19, 0.9830383658409119)
(20, 0.985988199710846)
(21, 0.966076672077179)
(22, 0.985988199710846)
(23, 0.982300877571106)
(24, 0.967551589012146)
(25, 0.98525071144104)
(26, 0.983775794506073)
(27, 0.9741888046264648)
(28, 0.98525071144104)
(29, 0.985988199710846)
(30, 0.98525071144104)
(31, 0.98525071144104)
(32, 0.98525071144104)
(33, 0.985988199710846)
(34, 0.983775794506073)
(35, 0.987463116645813)
(36, 0.9896755218505859)
(37, 0.9896755218505859)
(38, 0.98893803358078)
(39, 0.9896755218505859)
(40, 0.987463116645813)
(41, 0.9867256879806519)
(42, 0.987463116645813)
(43, 0.9882006049156189)
(44, 0.9867256879806519)
(45, 0.9867256879806519)
(46, 0.9867256879806519)
(47, 0.9867256879806519)
(48, 0.985988199710846)
(49, 0.987463116645813)
(50, 0.987463116645813)
(51, 0.9867256879806519)
(52, 0.987463116645813)
(53, 0.987463116645813)
(54, 0.9896755218505859)
(55, 0.987463116645813)
(56, 0.9882006049156189)
    };
\addlegendentry{16$\times$64}

\addplot[color = black, mark = circle]
    coordinates {
(0, 0.8230088353157043)
(1, 0.9233038425445557)
(2, 0.9756637215614319)
(3, 0.9653392434120178)
(4, 0.9417403936386108)
(5, 0.9594395160675049)
(6, 0.98893803358078)
(7, 0.985988199710846)
(8, 0.9882006049156189)
(9, 0.9904129505157471)
(10, 0.9292035102844238)
(11, 0.98893803358078)
(12, 0.98525071144104)
(13, 0.987463116645813)
(14, 0.9896755218505859)
(15, 0.9867256879806519)
(16, 0.991150438785553)
(17, 0.9933628439903259)
(18, 0.9830383658409119)
(19, 0.9830383658409119)
(20, 0.9955751895904541)
(21, 0.983775794506073)
(22, 0.99262535572052)
(23, 0.9896755218505859)
(24, 0.994837760925293)
(25, 0.994837760925293)
(26, 0.9955751895904541)
(27, 0.994837760925293)
(28, 0.9955751895904541)
(29, 0.9955751895904541)
(30, 0.9955751895904541)
(31, 0.9955751895904541)
(32, 0.9918879270553589)
(33, 0.9955751895904541)
(34, 0.994837760925293)
(35, 0.994837760925293)
(36, 0.991150438785553)
(37, 0.994837760925293)
(38, 0.994837760925293)
(39, 0.99262535572052)
(40, 0.99631267786026)
(41, 0.994837760925293)
(42, 0.994837760925293)
(43, 0.9941002726554871)
(44, 0.98893803358078)
(45, 0.994837760925293)
(46, 0.99631267786026)
(47, 0.994837760925293)
(48, 0.99631267786026)
(49, 0.99631267786026)
(50, 0.994837760925293)
(51, 0.9955751895904541)
(52, 0.9955751895904541)
(53, 0.99631267786026)
(54, 0.99631267786026)
(55, 0.99631267786026)
(56, 0.99631267786026)
(57, 0.99631267786026)
(58, 0.99631267786026)
(59, 0.99631267786026)
(60, 0.99631267786026)
    };
\addlegendentry{32$\times$32}

\addplot[color = cyan, mark = circle]
    coordinates {
(0, 0.8252212405204773)
(1, 0.9247787594795227)
(2, 0.9741888046264648)
(3, 0.966076672077179)
(4, 0.9550147652626038)
(5, 0.967551589012146)
(6, 0.980088472366333)
(7, 0.9867256879806519)
(8, 0.985988199710846)
(9, 0.987463116645813)
(10, 0.9262536764144897)
(11, 0.98893803358078)
(12, 0.9830383658409119)
(13, 0.987463116645813)
(14, 0.9867256879806519)
(15, 0.985988199710846)
(16, 0.98893803358078)
(17, 0.9904129505157471)
(18, 0.9867256879806519)
(19, 0.9830383658409119)
(20, 0.9904129505157471)
(21, 0.982300877571106)
(22, 0.98893803358078)
(23, 0.987463116645813)
(24, 0.9904129505157471)
(25, 0.987463116645813)
(26, 0.9904129505157471)
(27, 0.9904129505157471)
(28, 0.99262535572052)
(29, 0.9933628439903259)
(30, 0.9933628439903259)
(31, 0.98893803358078)
(32, 0.987463116645813)
(33, 0.9941002726554871)
(34, 0.9904129505157471)
(35, 0.9918879270553589)
(36, 0.98525071144104)
(37, 0.9904129505157471)
(38, 0.994837760925293)
(39, 0.9882006049156189)
(40, 0.9955751895904541)
(41, 0.9918879270553589)
(42, 0.9904129505157471)
(43, 0.991150438785553)
(44, 0.987463116645813)
(45, 0.9904129505157471)
(46, 0.9918879270553589)
(47, 0.9896755218505859)
(48, 0.991150438785553)
(49, 0.9933628439903259)
(50, 0.98893803358078)
(51, 0.9904129505157471)
(52, 0.9896755218505859)
(53, 0.991150438785553)
(54, 0.9941002726554871)
(55, 0.9904129505157471)
(56, 0.9933628439903259)
(57, 0.991150438785553)
(58, 0.9941002726554871)
(59, 0.9918879270553589)
(60, 0.9918879270553589)
    };
\addlegendentry{64$\times$16}

\addplot[color = orange, mark = circle]
    coordinates {
(0, 0.8694690465927124)
(1, 0.8643068075180054)
(2, 0.9137167930603027)
(3, 0.8672566413879395)
(4, 0.9446902871131897)
(5, 0.9476401209831238)
(6, 0.9432153105735779)
(7, 0.980088472366333)
(8, 0.9793510437011719)
(9, 0.974926233291626)
(10, 0.9638643264770508)
(11, 0.9719763994216919)
(12, 0.9756637215614319)
(13, 0.9646017551422119)
(14, 0.9830383658409119)
(15, 0.9815633893013)
(16, 0.9808259606361389)
(17, 0.983775794506073)
(18, 0.9867256879806519)
(19, 0.9793510437011719)
(20, 0.9808259606361389)
(21, 0.98893803358078)
(22, 0.98525071144104)
(23, 0.98893803358078)
(24, 0.985988199710846)
(25, 0.9808259606361389)
(26, 0.987463116645813)
(27, 0.9896755218505859)
(28, 0.985988199710846)
(29, 0.987463116645813)
(30, 0.9867256879806519)
(31, 0.9867256879806519)
(32, 0.98893803358078)
(33, 0.9867256879806519)
(34, 0.9882006049156189)
(35, 0.9867256879806519)
(36, 0.98525071144104)
(37, 0.9882006049156189)
(38, 0.9882006049156189)
(39, 0.9896755218505859)
(40, 0.985988199710846)
(41, 0.9882006049156189)
(42, 0.98893803358078)
(43, 0.98893803358078)
(44, 0.987463116645813)
(45, 0.9896755218505859)
(46, 0.9882006049156189)
(47, 0.98893803358078)
(48, 0.9882006049156189)
(49, 0.98893803358078)
(50, 0.9882006049156189)
(51, 0.987463116645813)
(52, 0.98893803358078)
(53, 0.987463116645813)
(54, 0.9882006049156189)
(55, 0.987463116645813)
(56, 0.98893803358078)
(57, 0.98893803358078)
(58, 0.98893803358078)
(59, 0.98893803358078)
    };
\addlegendentry{128$\times$8}

\addplot[color = purple, mark = circle]
    coordinates {
(0, 0.7544247508049011)
(1, 0.9461652040481567)
(2, 0.9351032376289368)
(3, 0.9151917099952698)
(4, 0.9616519212722778)
(5, 0.9631268382072449)
(6, 0.9616519212722778)
(7, 0.8709439635276794)
(8, 0.9439527988433838)
(9, 0.983775794506073)
(10, 0.974926233291626)
(11, 0.976401150226593)
(12, 0.976401150226593)
(13, 0.9734513163566589)
(14, 0.9815633893013)
(15, 0.9808259606361389)
(16, 0.98525071144104)
(17, 0.98525071144104)
(18, 0.9845132827758789)
(19, 0.983775794506073)
(20, 0.9815633893013)
(21, 0.985988199710846)
(22, 0.983775794506073)
(23, 0.98893803358078)
(24, 0.980088472366333)
(25, 0.991150438785553)
(26, 0.976401150226593)
(27, 0.991150438785553)
(28, 0.9896755218505859)
(29, 0.9904129505157471)
(30, 0.9904129505157471)
(31, 0.9904129505157471)
(32, 0.98893803358078)
(33, 0.98893803358078)
(34, 0.985988199710846)
(35, 0.9896755218505859)
(36, 0.991150438785553)
(37, 0.987463116645813)
(38, 0.991150438785553)
(39, 0.9867256879806519)
(40, 0.9904129505157471)
(41, 0.9904129505157471)
(42, 0.991150438785553)
(43, 0.9904129505157471)
(44, 0.991150438785553)
(45, 0.9918879270553589)
(46, 0.9896755218505859)
(47, 0.9904129505157471)
(48, 0.9904129505157471)
(49, 0.991150438785553)
(50, 0.991150438785553)
(51, 0.9918879270553589)
(52, 0.9918879270553589)
(53, 0.9904129505157471)
(54, 0.9918879270553589)
(55, 0.9904129505157471)
(56, 0.9918879270553589)
(57, 0.991150438785553)
(58, 0.9918879270553589)
(59, 0.991150438785553)
(60, 0.991150438785553)
(61, 0.991150438785553)
(62, 0.9918879270553589)
    };
\addlegendentry{1024$\times$1}
\end{axis}
\end{tikzpicture}